\documentclass[10pt,twoside]{article}
\usepackage{amsmath}
\usepackage[utf8]{inputenc}
\usepackage{url}
\usepackage{dsfont}
\usepackage{graphicx}
\usepackage{yhmath}
\usepackage[table]{xcolor}
\usepackage{mathrsfs} 
\usepackage{amssymb}
\usepackage{url}
\usepackage{makecell}
\usepackage{arydshln}
\usepackage{multirow}
\usepackage{arydshln}
\usepackage{mathtools}
\usepackage{stmaryrd}  
 \usepackage{amsthm,amssymb}

\input epsf
\def\limiten{\renewcommand{\arraystretch}{0.5}
\begin{array}[t]{c}\stackrel{}{\longrightarrow} \\
{\scriptstyle n\rightarrow
\infty}\end{array}\renewcommand{\arraystretch}{1}}

\numberwithin{equation}{section}

\newtheorem{thm}{Theorem}[section]

\newtheorem{Def}[thm]{Definition\rm}

\newtheorem{rmrk}[thm]{Remark}

\newcommand{\E}{\ensuremath{\mathbb{E}}}
\newcommand{\R}{\ensuremath{\mathbb{R}}}
\newcommand{\Z}{\ensuremath{\mathbb{Z}}}

\newcommand{\N}{\ensuremath{\mathbb{N}}}

\newcommand{\cov}{\ensuremath{\mathrm{Cov}}}

\newcommand{\lip}{\ensuremath{\mathrm{Lip}}}
\definecolor{grisclair}{gray}{0.9}
\font\dsrom=dsrom10 scaled 1200
\def \ind{\textrm{\dsrom{1}}}
\DeclareMathOperator*{\argmin}{argmin}

\renewcommand{\arraystretch}{.8}

\begin{document}

\title{\bf Sparse-penalized deep neural networks estimator  under weak dependence}
 \maketitle \vspace{-1.0cm}

\begin{center}
      William Kengne \footnote{Developed within the ANR BREAKRISK: ANR-17-CE26-0001-01 and the  CY Initiative of Excellence (grant "Investissements d'Avenir" ANR-16-IDEX-0008), Project "EcoDep" PSI-AAP2020-0000000013
   } 
   and 
     Modou Wade \footnote{Supported by the MME-DII center of excellence (ANR-11-LABEX-0023-01)} 
 \end{center}

  \begin{center}
  { \it 
 THEMA, CY Cergy Paris Université, 33 Boulevard du Port, 95011 Cergy-Pontoise Cedex, France\\
  E-mail:   william.kengne@cyu.fr  ; modou.wade@cyu.fr\\
  }
\end{center}

 \pagestyle{myheadings}

\markboth{Sparse-Penalized deep neural networks estimator under weak dependence}{Kengne and Wade}

\medskip

\textbf{Abstract}:
We consider the nonparametric regression and the classification problems for $\psi$-weakly dependent processes.  
This weak dependence structure is more general than conditions such as, mixing, association, $\ldots$.
A penalized estimation method for sparse deep neural networks is performed.
In both nonparametric regression and binary classification problems, we establish oracle inequalities for the excess risk of the sparse-penalized deep neural networks estimators.
Convergence rates of the excess risk of these estimators are also derived. 
The simulation results displayed show that, the proposed estimators overall work well than the non penalized estimators.

\medskip
 

\medskip

\section{Introduction}
Deep learning has received a considerable  attention in the literature and has shown great success in several applications  in  artificial intelligence, for example,  image processing (see (\cite{krizhevsky2017imagenet})),  speech recognition (see(\cite{hinton2012deep})) $\cdots$.
In  the recent decades, many researchers have contributed to understand  the theoretical properties of deep neural networks (DNNs) predictors with  the sparse regularization.
See, for instance, \cite{ohn2019smooth}, \cite{schmidt2019deep}, \cite{bauer2019deep}, \cite{schmidt2020nonparametric}, \cite{tsuji2021estimation}, \cite{kim2021fast} \cite{imaizumi2022advantage} (and the references therein) for some results with independent and identical distribution (i.i.d.) observations, and \cite{chen2019bbs}, \cite{kohler2020rate},  \cite{kurisu2022adaptive}, \cite{ma2022theoretical}, \cite{kengne2023deep}, \cite{kengne2023excess} for some results with dependent or non-i.i.d. observations. 
Sparse-penalized DNNs estimators have been studied, among others, by \cite{ohn2022nonconvex}, \cite{kurisu2022adaptive}.
These authors establish some oracle-type inequalities for i.i.d. data and dependent observations under a mixing condition.

\medskip

Let us  consider the set of observations $ D_n \coloneqq  \{ (X_1, Y_1), \cdots, (X_n, Y_n) \} $ (the training sample) from a stationary and ergodic process $\{Z_t =(X_t, Y_t), t \in \Z \} $, which takes values in $\mathcal{Z}=(\mathcal{X}  \times \mathcal{Y}) $,  where $\mathcal{X}$ is the input space and $\mathcal{Y}$ the output space.
Consider for the sequel, a loss function  $\ell: \R \times  \mathcal{Y} \rightarrow [0,\infty) $.

\medskip

\noindent We perform sparse DNNs predictors for nonparametric regression  and classification tasks, based on a penalized empirical risk minimization procedure.
The estimator $\widehat{h}_n \in \mathcal{H}_{\sigma}(L_n, N_n, B_n, F) $ is also called, the sparse-penalized DNN (SPDNN) predictor, and is given by 
\begin{equation}\label{sparse_DNNs_Estimators}
\widehat{h}_n = \underset{h \in \mathcal{H}_{\sigma}(L_n, N_n, B_n, F)}{\argmin} \left[ \dfrac{1}{n} \sum_{i=1}^{n} \ell(h(X_i), Y_i) + J_n(h) \right],
\end{equation}
where $\mathcal{H}_{\sigma}(L_n, N_n, B_n, F) $ is a class of DNNs (see (\ref{DNNs_no_Constraint})) with an activation function $\sigma$ and with suitably chosen architecture parameters $L_n, N_n, B_n, F$; and
 $J_ {n} (h) $ is the sparse penalty given by 
\[ J_ {n} (h) \coloneqq J_ {\lambda_ {n}, \tau_ {n}} (h) \coloneqq \lambda_ {n} \|\theta(h) \|_ {\text{clip}, \tau_ {n}},  \]  
for tuning parameters $\lambda_ {n} > 0, \; \tau_ {n} > 0$ and $\theta(h)$ is the vector of parameters of $h$. 
$\| \cdot \|_{\text{clip}, \tau} $ denotes the clipped $ L_1$ norm  with a 
  clipping threshold $ \tau > 0$ see \cite{zhang2010analysis} defined as 
\[ \| \theta \|_{\text{clip}, \tau} =  \sum_{j=1}^{p}\left(\frac{|\theta_{j}|}{\tau}\land 
  1\right), \] 
where $\theta= (\theta_1, \cdots, \theta_ p)' $ is a  $p$-dimensional 
  vector and $'$ denotes the transpose. 

\medskip

For both nonparametric regression and classification under weak dependence, 
we focus on the calibration of the parameters $L_n, N_n, B_n, F$ of the networks class, and $\lambda_n$, $\tau_n$ of the penalty term, that allow the SPDNN predictor $\widehat{h}_n$ to enjoy an oracle property and to derive its convergence rate.
These issues have been addressed by \cite{ohn2022nonconvex} in the i.i.d. case.
They have established an oracle inequality of the excess risk 
and have proved that, the SPDNN can adaptively attain minimax optimality.
\cite{kurisu2022adaptive} carried out SPDNN estimator for nonparametric time series regression under $\beta$-mixing condition.
They provided a generalization error bound of the SPDNN estimator and proved that, this estimator attains the minimax optimal rate up to a poly-logarithmic factor.
Besides addressing the time series classification issue, the $\psi$-weak dependence considered here is more general than mixing conditions (see \cite{dedecker2007weak}).

\medskip

In this new contribution, we consider SPDNNs estimators for learning a $\psi$-weakly dependent process $ \{Z_t = (X_t, Y_t), ~ t  \in \Z \} $ with value in $ Z= \mathcal{X} \times \mathcal{Y} \subseteq \R^d \times \R $, based on the training sample $D_n = {(X_1, Y_1), \cdots, (X_n, Y_n)} $ and we address the following issues.

\begin{itemize}
\item [(i)] \textbf{Oracle inequality for the nonparametric time series regression}. 
 We provide conditions on  $L_n, N_n, B_n, F, \lambda_n$, $\tau_n$, and establish an oracle inequality of the $L_2 $ error of the SPDNN estimator.
\item[(ii)] \textbf{Oracle inequality for the binary time series classification}.
Conditions on  $L_n, N_n, B_n, F, \lambda_n$, $\tau_n$ are provided, and an oracle inequality of the excess risk of the SPDNN estimator is established. 
\item[(iii)] \textbf{Convergence rates of the excess risk}. 
  For both nonparametric regression and time series classification, the convergence rate (which depends on $L_n, N_n, B_n, F, \lambda_n, \tau_n$) of the excess risk is derived. When the true regression function (in the regression problem) and the target function (in the classification task) are sufficiently smooth, these rates are close to $\mathcal{O}(n^{-1/2})$.  
\end{itemize}

\medskip

The rest of the paper is organized as follows. In Section \ref{asump}, we set some  notations and assumptions.  
Section \ref{Def_DNNs} defines the class of DNNs considered.
Section \ref{regression} is devoted to the nonparametric regression whereas Section \ref{classification} focuses on the binary time series classification. 
Some simulation results are provided in Section \ref{some_sim} and  Section \ref{prove} is devoted to the proof of the main results.

\medskip

\section{Notations and assumptions}\label{asump}
For  two separable Banach spaces $E_1, E_2$  equipped with norms $\| \cdot\|_{E_1} $ and $\| \cdot\|_{E_2} $ respectively, denote by $\mathcal{F}(E_1, E_2) $, the set of measurable functions from $E_1$ to $E_2$. For any $h \in \mathcal{F}(E_1, E_2) $ and $\epsilon >0$, $B(h,\epsilon) $ denotes the ball of radius $\epsilon$ of $\mathcal{F}(E_1, E_2) $ centered at $h$, that is,
\[ B (h, \epsilon) = \big\{ f \in \mathcal{F}(E_1, E_2), ~ \| f - h\|_\infty \leq \epsilon \big\},  \]
where $\| \cdot \|_\infty$ stands for the sup-norm defined below.
Let $\mathcal{H} \subset \mathcal{F}(E_1, E_2) $, the $\epsilon$-covering number $\mathcal{N}(\mathcal{H},\epsilon) $ of $\mathcal{H} $ is   the minimal 
  number of balls of radius $\epsilon$ needed to cover  $\mathcal{H} $; that is,
\[ \mathcal{N}(\mathcal{H},\epsilon) = \inf\Big\{ m \geq 1 ~ : \exists h_1, \cdots, h_m \in \mathcal{H} ~ \text{such that} ~ \mathcal{H} \subset 
  \bigcup_{i=1}^m B(h_i,\epsilon)    \Big\}.  \]
 For a function $h: E_1 \rightarrow E_2$ and $U \subseteq E_1$, define,
\[ \| h\|_\infty = \sup_{x \in E_1} \| h(x) \|_{E_2}, ~ \| h\|_{\infty,U} = \sup_{x \in U} \| h(x) \|_{E_2} ~ \text{and}\] 
\[\lip_\alpha (h) \coloneqq \underset{x_1, x_2 \in E_1, ~ x_1\ne x_2}{\sup} \dfrac{ \|h(x_1) - h(x_2)\|_{E_2}}{\| x_1- x_2 \|^\alpha_{E_1}} 
~ \text{for any}  ~ \alpha \in [0,1]. \]
 For any $\mathcal{K}_{\ell} > 0$ and $\alpha \in [0,1]$, $\Lambda_{\alpha,\mathcal{K}_{\ell}} (E_1, E_2) $ (simply $\Lambda_{\alpha,\mathcal{K}_{\ell}} (E_1) $ when $E_2 \subseteq \R$) is the set of functions  $h: E_1^u  \rightarrow E_2$ for some $u \in \N$, satisfies  $\|h\|_\infty < \infty$ and  $\lip_\alpha(h) \leq \mathcal{K}_{\ell}$.
When $\alpha=1$, we set  $\lip_1 (h)=\lip(h)$ and $\Lambda_{1}(E_1) =\Lambda_{1,1}(E_1,\R) $. 

\medskip

We define the weak dependence structure, see \cite{doukhan1999new} and \cite{dedecker2007weak}. Let $E$ be a separable Banach space.

\begin{Def}\label{Def}
$ An\; E- valued \;process \;(Z_t)_{t \in \Z}\; is\; said\; to\; be\; (\Lambda_1(E), 
  \psi,\epsilon)- weakly\; dependent\; $  $ if\; there\; exists\; a\;\\ function\; $ $\psi:[0, \infty)^2 \times \N^2 \to [0, \infty) \;and\; $ $a\; sequence\; \epsilon=(\epsilon(r))_{r \in \N}\; decreasing\; to\; zero\; at\; infinity\; such\; that,\; for\;\\ any\; g_1, \;g_2 \in \Lambda_1 (E)\; with\; g_1: E^u \rightarrow \R, \;g_2: E^v \rightarrow \R\;, ~ (u, v \in \N)\; and\; for\; any\;  u-tuple\; (s_1, \cdots, s_u)\; and\; any\; v-tuple\; (t_1, \cdots, t_v)\; with\; s_1 \leq \cdots \leq s_u \leq s_u + r \leq t_1 \leq \cdots \leq t_v, \; the\; following\; inequality\; is\; fulfilled$:
\[ \vert \cov (g_{1}(Z_{s_1}, \cdots, Z_{s_u}),  g_{2}(Z_{t_1}, \cdots, 
  Z_{t_v})) \vert \leq \psi(\lip(g_1),\lip(g_2), u, v) \epsilon(r). \]
\end{Def}
For example, following choices of $\psi$ leads to some well-known weak dependence conditions. 
\begin{itemize}
\item $\psi \left(\lip(g_1),\lip(g_2), u, v \right) = v \lip(g_2) $: the $\theta$-weak dependence, then denote $\epsilon(r) = \theta(r) $;
\item $\psi \left(\lip(g_1),\lip(g_2),u,v \right)= u \lip(g_1) + v \lip(g_2)$: the $\eta$-weak dependence, then denote $\epsilon(r) = 
 \eta(r) $;
\item $\psi \left(\lip(g_1),\lip(g_2), u, v \right)= u v \lip(g_1) \cdot \lip(g_2) $: the $\kappa$- weak dependence, then denote $\epsilon(r) = \kappa(r) $;
\item $\psi \left(\lip(g_1), \lip(g_2), u, v \right) = u \lip(g_1) + v \lip(g_2) + u v \lip(g_1) \cdot \lip(g_2) $: the $\lambda$-weak dependence, then denote $\epsilon(r) = \lambda(r) $. 
\end{itemize}
\medskip
 
We consider the  process $\{ Z_t=(X_t, Y_t), t \in \Z \}$ with  values 
   in $ \mathcal{Z}=   \mathcal{X} \times \mathcal{Y} \subset \R^d \times \R$, the loss function $\ell:\R \times \mathcal{Y} \rightarrow [0, \infty)$, the class of DNNs $\mathcal{H}_{\sigma}(L_n, N_n, B_n, F) $ with the activation function $\sigma: \R \rightarrow \R $, and set of following  assumptions.
\begin{itemize}
\item[\textbf{(A1)}]: There exists  a constant $C_{\sigma} > 0$ such that the 
   activation function $\sigma \in \Lambda_{1, C_{\sigma}}(\R) $.
\item[\textbf{(A2)}]: There exists $\mathcal{K}_{\ell} > 0$ such that, the loss 
  function $\ell \in \Lambda_{1, \mathcal{K}_{\ell}}(\R \times \mathcal{Y})$ and $M={\sup_{h\in \mathcal{H}_{\sigma}(L_n, N_n, B_n, F)}}{\sup_{z \in \mathcal{Z}}|\ell(h, z) |} < \infty$. 
  In the case of margin-based loss, these conditions are assumed to the function $(u, y) \mapsto \ell(uy)$.
 \end{itemize}
Let us set the weak dependence assumption.

\begin{itemize}
\item[\textbf{(A3)}]: The process $\left\{ Z_t = (X_t, Y_t), {t \in \Z} \right\}$  is stationary ergodic and   $(\Lambda_1 (\mathcal{Z}), \psi, \epsilon)$-weakly dependent with $\epsilon_r = \mathcal{O}(r^{-\gamma})$ 
  for some $\gamma > 3$.
\end{itemize}

\section{ Deep Neural Networks }\label{Def_DNNs}
A DNN, with $ (L,   \textbf{p}) $  network architecture, where $L \in \N$ is the number of hidden-layers and $ \textbf{p}=(p_{0},   \cdots,  p_{L+1}) \in \N^{L+2}$ the width vector, is any function  $h$ of the form, 
\begin{equation}\label{h_equ1}
h: \R^{p_0} \rightarrow \R^{p_{L+1}}, \;   x\mapsto h(x) = A_{L+1} \circ \sigma_{L} \circ A_{L} \circ \sigma_{L-1} \circ \cdots \circ \sigma_1 \circ A_1 (x),
\end{equation} 
where $ A_j: \R^ {p_ {j -1}} \rightarrow \R^ {p_j} $ is a linear affine map, defined by $A_j (x) \coloneqq W_j x + \textbf{b}_j$,  for given  $p_ {j - 1}\times p_j$  weight matrix   $ W_j$   and a shift vector $ \textbf{b}_j \in \R^ {p_j} $  and $\sigma_j: \R^{p_j} \rightarrow \R^ {p_j} $ is a nonlinear element-wise activation map, defined by $\sigma_j (z) = (\sigma(z_1), \cdots, \sigma(z_{p_j}))^ {'} $.
For a DNN of the form (\ref{h_equ1}), denote by,
\begin{equation} \label{def_theta_h}
 \theta(h) \coloneqq \left(vec(W_1)^ {'}, \textbf{b}^{'}_{1}, \cdots,  vec(W_{L + 1})^{'} , \textbf{b}^ {'}_{L+1} \right)^{'}, 
 \end{equation} 
the vector of its parameters, where $ vec(W)$ transforms the matrix $W$ into the corresponding vector by concatenating the  column vectors. 
Let $\mathcal{H}_{\sigma, p_0, p_ {L+1}} $ be the class of DNNs predictors that take $p_0$-dimensional input to produce $p_ {L+1} $-dimensional output and use the activation function $ \sigma: \R \rightarrow \R$.
In our setting here, $p_0 = d$ and $p_ {L + 1} = 1$.
For a DNN $h$ as in (\ref{h_equ1}), let depth($h$) and width($h$) be  respectively the depth and width of $h$; that is, depth($h$)$=L$ and width($h$) = $\underset{1\leq j \leq L} {\max} p_j $. For any positive constants $L, N,   B$ and $F$, we set
\[ \mathcal{H}_{\sigma}(L, N, B) \coloneqq \big \{h\in \mathcal{H}_{\sigma, q, 1}: \text{depth}(h)\leq L, \text{width}(h)\leq N, \|\theta(h)\|_{\infty} \leq B \big\},  \]
 and
\begin{equation}\label{DNNs_no_Constraint}
\mathcal{H}_{\sigma}(L, N, B, F) \coloneqq \big\{ h: h\in H_{\sigma}(L, N, B), \| h \|_{\infty, \mathcal{X}} \leq F \big\}. 
\end{equation}
A class of sparsity constrained DNNs with sparsity level $S > 0$  is defined by 
\begin{equation}\label{DNNs_Constraint}
\mathcal{H}_{\sigma}(L, N, B, F, S) \coloneqq \left\{h\in \mathcal{H}_{\sigma}(L, N, B, F) \; :  \; \| \theta(h) \|_0 \leq S 
  \right\},
\end{equation}
where $\| x \|_0 = \sum_{i=1}^p \ind(x_i \neq 0)$ for all $x=(x_1,\ldots,x_p)' \in \R^p$ ($p \in \N$).
In the sequel, we  will establish some theoretical results for the SPDNN estimators (\ref{sparse_DNNs_Estimators}) in both regression and classification problems under the $\psi$-weak dependence. 
\section{Nonparametric regression}\label{regression}
In this section we aim to study the nonparametric time series regression, with the output $ Y_t \in \R$ and the input $ X_t \in \R^d$ are generated from the model,
\begin{equation}\label{regression_model}
Y_t = h^ {*} (X_t) + \epsilon_t, ~ X_0 \sim  \text{P}_{X_0},
\end{equation}
where $ h^ {*}: \R^d  \rightarrow \R$  is the  unknown regression  function and $\epsilon_t$ is an error variable independent of the input variable  $X_t$.  Let us consider the  sub-Gaussian assumption to the error, 
\begin{equation}\label{sub_Gaussian}
\E [ e^ {t\epsilon_t}]  \leq e^ {t^2 \rho^2 / 2},
\end{equation}
for any $ t \in \R$ for some $\rho > 0$. 
So, denote by $ \mathcal{H}_{\rho, H^ {*}} $ the set of distributions $ (X_0, Y_0) $ satisfying the model (\ref{regression_model})
\begin{equation}\label{def_H_rho_H_star}
\mathcal{H}_{\rho, H^ {*}} \coloneqq \left\{ \text{Model}  (\ref{regression_model}): \E [ e^{t\epsilon_t}]  \leq e^{t^2\rho^2 / 2}, \forall t \in \R, \|h^ {*}\|_{\infty} \leq H^ {*} \right\}. 
\end{equation}
We will focus on the estimate of the unknown regression function $h^ {*} $ based on the  training sample $\{(X_i, Y_i)\}_{1 \leq i \leq n}$, with $ (X_0,Y_0) \sim  H $ where $ H \in \mathcal{H}_{\rho, H^ {*}} $.
We consider the $L_2$ error as the measure of the performance of the  estimator $ \widehat{h}_n$,
\[ \E \left[\| \widehat{h}_n - h^ {*}\|_{2, P_ {X_0}} ^2 \right],   \]
where
\[ \| \widehat{h}_n - h^ {*}\|_{r, P_{X_0}}^r  \coloneqq \displaystyle \int \vert  \widehat{h}_n (\text{x})- h^ {*} (\text{x}) \vert^r  d P_{X_0} ( \text{x}),   \]
for all $r \geq 1$.

\medskip

In the sequel, we use the notations, $a \lor b = \max(a,b)$, 
$ a_n \lesssim b_n$ or  $ b_n \gtrsim  a_n$  if there exists a positive constant $C> 0$ such that $ a_n \leq  C b_n$ for all $ n \in \N$, $a_n \asymp b_n$ if $ a_n \lesssim b_n$ and $ a_n \gtrsim b_n$.
The following Theorem provides an oracle inequality of the $ L_2$ error of the SPDNN estimator.

\begin{thm}\label{thm1}
Assume (\textbf{A1})-(\textbf{A3}) and that, $(X_0, Y_0) \sim H \in \mathcal{H}_{\rho, H^ {*}} $. Let  $F > 0$,
  $L_n \lesssim \log n, N_n \lesssim n^{\nu_1},  1 \leq  B_n \lesssim n^{\nu_2} $ for some $\nu_1 >0, ~ \nu_2 > 0$. Then, the sparse-penalized DNN estimator defined by 
\begin{equation}\label{EMR_algo}
\widehat{h}_n = \underset{h\in \mathcal{H}_{\sigma}(L_n, N_n, B_n, F)}{\argmin}\Big[\dfrac{1}{n} \sum_{i=1} ^{n} \left(Y_i - h(X_i)\right) ^{2} + \lambda_n \|\theta (h) \|_{ \text{clip}, \tau_n} \Big],  
\end{equation}
with $ \lambda_n \asymp (\log n) ^{\nu_3} /n^{\nu_4}, ~ \nu_3 > 0, ~  0< \nu_4 < 1, ~ \tau_n \leq  
  \dfrac{\beta_n}{16 K_n  (L_n + 1)((N_n + 1) B_n)^{L_n +1}} ~ \text{for} ~ \beta_n \coloneqq (\log n)^{\nu_5} /n^{\nu_6} $, satisfies
\begin{align}\label{bound_error}
\nonumber & \E\Big[ \| \widehat{h}_n - h^ {*} \|_{2, P_ {X_0}} ^2 \Big]  
\\
& \leq 2\Bigg( \underset{h \in \mathcal{H}_{\sigma}(L_n, N_n, B_n,  F)}{\inf} \{ \| h - h^ {*} \|_{2, P_ {X_0}}^2 + \lambda_n \|\theta (h)\|_ { \text{clip}, \tau_n} \} \lor  \dfrac{C_{ \sigma, H^ {*}} (\log n)^{\nu_5}}{n^{\nu_6}}  \Bigg);
\end{align}
for some constant $C_ { \rho, H^ {*}} > 0$ depending on  $\rho$ and $H^ {*}, \; \text{with} \;  
  \nu_5 >0, ~ \nu_6 >0, ~ \Big( \nu_6 < 1/2, \nu_4 + \nu_6 <1  \Big) \text{ or } \Big( \nu_6 < 1/2, \nu_4 + \nu_6 =1, \nu_5 > 1-\nu_3 \Big)$.
\end{thm}

\medskip

 The next theorem provides a useful tool to derive convergence rates of the SPDNN estimator in several situations. 
\begin{thm}\label{thm2}
Assume the conditions of Theorem \ref{thm1}, with similar choices of $F, L_n, N_n, B_n, \lambda_n, \tau_n$.
Moreover, let $H^ {*} > 0$  and $\mathcal{H}^ {*} $ be a set of real real-value function on $ \R^d$ and assume there are constants $\kappa > 0, ~ r > 0, \epsilon_0 > 0$, and $C > 0$ such that, 
\begin{equation}\label{equ_assump_thm2}
\underset{ h^ {\diamond} \in \mathcal{H}^ {*}: \|h^ {\diamond} \|_ {\infty} \leq H^ {*} }{\sup} ~ \underset{ h \in  \mathcal{H}_ {\sigma} (L_n, N_n, B_n, F, S_ {n, \epsilon})}{\inf} \|h - h^ {\diamond} \|_{2, P_{X_0}} \leq \epsilon,
\end{equation}
with $S_ {n, \epsilon} \coloneqq C \epsilon^ {-\kappa} (\log n) ^r$ for any $\epsilon \in  (0, \epsilon_0) $ and $ n\in  \N$. Then, the SPDNN estimator defined in (\ref{EMR_algo})  satisfies
\begin{equation}\label{EMR_bound}
\underset{H\in \mathcal{H}_ {\rho, H^ {*}}: h^ {*} \in \mathcal{H}^ {*}} {\sup}  \E \Big[ \| \widehat{h}_n - h^ {*} \|_{2, P_{X_0}}^2 \Big] \lesssim \dfrac{ (\log n) ^{r + \nu_3}}{n^{ \frac{2 \nu_4}{\kappa + 2}} }  \lor \dfrac{(\log n) ^{\nu_5}}{n^ {\nu_6}},
\end{equation}
for some constant $\nu_3 > 0, ~ 0< \nu_4 < 1, ~ \nu_5 > 0, ~ \nu_6 > 0 ~ \Big( \nu_6 < 1/2, \nu_4 + \nu_6 < 1  \Big) ~ \text{or} ~\Big( \nu_6 < 1/2, \nu_4 + \nu_6 = 1, \nu_5 > 1-\nu_3 \Big) $.
\end{thm}

\begin{rmrk}\label{remark_rg}
 Consider the H{\"o}lder space of smoothness $s>0$, with radius $\mathcal{K}>0$ given by,
\begin{equation*}
 \mathcal{C}^{s,\mathcal{K}}( \mathcal{X} ) = \big\{h : \mathcal{X} \rightarrow \R,  ~  \| h\|_{\mathcal{C}^s(\mathcal{X} )} \leq  \mathcal{K} \big\},
\end{equation*}
where  $ \|  \cdot \|_{\mathcal{C}^s(\mathcal{X} )} $ denotes the H{\"o}lder norm defined by,
\[ \|h \|_{\mathcal{C}^s(\mathcal{X})} = \sum_{\beta \in \N_0^d, ~ |\beta| \leq [s]} \| \partial^\beta h \|_\infty + \sum_{\beta \in \N_0^d, ~ |\beta| = [s]} Lip_{s-[s]}(\partial^\beta h), \]
with, $\N_0 = \N \cup \{0\}$, $|\beta| = \sum_{i=1}^d \beta_i$ for all $\beta=(\beta_1,\ldots,\beta_d)' \in \N_0^d$, $\partial^\beta$ denotes the partial derivative of order $\beta$ and $[x]$ denotes the integer part of $x$.
So, if the true regression function satisfies for instance $h^* \in  \mathcal{C}^{s,\mathcal{K}}( \mathcal{X} )$ for some $\mathcal{K} >0$, then, condition (\ref{equ_assump_thm2}) holds for a  various classes of activation function, including ReLU with $\kappa=d/s$ and $r=1$ (see \cite{kengne2023deep}). In this case, if $s\gg d$, then, the convergence rate of the SPDNN estimator is close to $\mathcal{O}(n^{-1/2})$. 
\end{rmrk}

\section{Binary time series classification}\label{classification}
We consider the binary classification of the process $\{Z_t =(X_t, Y_t), t \in \Z \} $, which values in $\R^d \times \{-1, 1\}$. 
The goal is to construct a function $h: \R^d \rightarrow \R$ such that, $ h (X_t) $ is used to predict the label $ Y_t \in \{-1, 1\} $.
We will focus on a  margin-based loss function  to evaluate  the performance of  the prediction   by $h$;
;
that is, a loss function of the form: $(u,y) \mapsto \ell(uy)$. For instance, the hinge loss with $\ell(uy) = \max(1-uy, 0)$. 
 We assume in the sequel that, the input $ X_t \in \R^d$ and  the output $ Y_t  \in  \{-1, 1 \} $ are generated from the model 
 \begin{equation}\label{classifier_model}
Y_t  \vert X_t = \text{x}  \sim 2 \mathcal{B} (\eta ( \text{x})) - 1,  \qquad X_0 \sim \text{P}_ {X_0},
 \end{equation}
where $\eta( \text{x}) = P (Y_t = 1 \vert X_t = \text{x}) $ and $\mathcal{B} (\eta (\text{x})) $ is the Bernoulli distribution with parameter $ \eta (\text{x}) $.
The aim is to build a classifier $h$ so that, the excess risk of $h$ defined by,
\begin{equation}\label{def_excess_risk}
\mathcal{E}_{Z_0} (h) \coloneqq \E [\ell (Y_0  h(X_0))] - \E [\ell(Y_0 h_ {\ell}^ {*} (X_0))], ~ \text{with}~ Z_0 \coloneqq (X_0, Y_0)
\end{equation} 
is "close" to zero, where $\ell$ is a given margin-based loss 
  function, $h_ {\ell}^ {*} = \underset{h \in \mathcal{F}}{\argmin} \E \left[ \ell (Y_0 h (X_0))  \right]$ is a target function, suppose to be bound, that is $ \| h_ { \ell}^ {*} \|_{\infty} \leq H^ {*}$ for some $H^ {*} > 0$ and $\mathcal{F} $ is the set of measurable functions from $\R^d$ to $\R$.
For the sequel, define the set of distributions satisfying this assumption by, 
\begin{equation}
\mathcal{ Q}_ { H^ {*}} \coloneqq \{ Model (\ref{classifier_model}):  \| h_{ \ell}^ {*} \|_{\infty} \leq H^ {*} \}.
\end{equation}
The following theorem provides an oracle inequality for the excess risk of the SPDNN estimator based on a margin-based loss function $\ell$; whose, the strict convexity is not required as in \cite{ohn2022nonconvex}.
\begin{thm}\label{thm3}
Assume that (\textbf{A1})-(\textbf{A3}) hold and that the true generative model $H$ is in $\mathcal{Q}_ {H^ {*}} $.
Let  $F > 0$, $L_n \lesssim \log n, N_n \lesssim n^{\nu_1},  1 \leq  B_n \lesssim n^{\nu_2} $ for some $\nu_1 >0, ~ \nu_2 > 0$.
Then, the SPDNN estimator defined by, 
\begin{equation}\label{def_SPDNN_class}
\widehat{h}_n = \underset{h \in \mathcal{H}_ {\sigma}(L_n, N_n, B_n, F) }{\argmin}\left[ \dfrac{1}{n} \sum_{i=1}^{n} \ell(Y_i h (X_i)) + \lambda_n \| \theta (h)\|_{\text{clip}, \tau_n}  \right],
\end{equation}
with $\lambda_n \asymp (\log n) ^{\nu_3} /n, ~ \tau_n \leq  \dfrac{\beta_n}{4 \mathcal{K}_ {\ell}  (L_n + 1)((N_n + 1) B_n) ^{L_n +1}} ~ 
\text{for} ~ \beta_n \coloneqq (\log n) ^{\nu_5} /n^ {\nu_6}, ~  \mathcal{K}_ {\ell} > 0$, satisfies
\begin{equation}
\E \left[ \mathcal{E}_{Z_0} (\widehat{h}_n)\right] \leq 2 \underset{h \in \mathcal{H}_ {\sigma}( L_n, N_n, B_n, F)}{\inf} \left\{ \mathcal{E}_{Z_0} (h) +  \lambda_n \|\theta (h)\|_{\text{clip}, \tau_n} \right\} \lor \dfrac{C (\log n) ^{\nu_5} }{ n^ {\nu_6}},
\end{equation}
for some  universal constant  $ C >0, ~ \nu_5 > 0, ~ \nu_6 > 0, ~ \Big( \nu_6 < 1/2, \nu_4 + \nu_6 < 1  \Big) ~ \text{or} ~ \Big( 
\nu_6 < 1/2, \nu_4 + \nu_6 = 1, \nu_5 > 1-\nu_3 \Big) $, where the expectation is taken over the training data.
 \end{thm}

As Theorem \ref{thm2} in the nonparametric regression, the following theorem provides a useful tool to derive a convergence rate of the SPDNN estimator.

\begin{thm}\label{thm4}
Assume the conditions of Theorem \ref{thm3}, with similar choices of $F, L_n, N_n, B_n, \lambda_n, \tau_n$. 
Let $H^{*} > 0$ and let $\mathcal{H}^ {*} $ be a set of real-valued functions on $\R^d$  and assume there are constants $\kappa > 0, ~ r > 0, \epsilon_0 > 0$, and $C > 0$  such that, 
\begin{equation}\label{equ_assump_thm4}
\underset{ h^\diamond \in \mathcal{H}^ {*}: \|h^\diamond \|_{\infty} \leq H^ {*} }{\sup} ~\underset{ h \in  \mathcal{H}_\sigma (L_n, N_n, B_n, F, S_{n, \epsilon})}{\inf} \| h - h^\diamond \|_{1, P_{X_0}} \leq \epsilon,
\end{equation}
with $ S_ {n, \epsilon} \coloneqq C \epsilon^ {-\kappa} (\log n) ^r $ for any $ 
\epsilon \in  (0, \epsilon_0) $ and $ n \in  \N$. Then, SPDNN estimator defined in (\ref{def_SPDNN_class}) satisfies, 
\begin{equation}\label{bound_exces_risk}
\underset{H \in \mathcal{Q}_ {H^ {*}}: h_{\ell}^ {*} \in \mathcal{H}^ {*} }{\sup} \E \left[ \mathcal{E}_{Z_0} (\widehat{h}_n) \right] \lesssim \dfrac{ (\log n)^ {r + \nu_3} }{n^ { \frac{\nu_4}{\kappa + 1}  }}  \lor \dfrac{ (\log n) ^{\nu_5} }{n^{ \nu_6}}.
\end{equation}
   
\end{thm}
\noindent As stressed in Remark \ref{remark_rg}, such result can be used to derive a convergence rate of the excess risk of the SPDNN estimator in several situations.

\medskip

\section{Some numerical results}\label{some_sim}
In this section, we carry out the prediction of autoregressive models and binary time series by SPDNN.
\subsection{Prediction of autoregressive models}\label{Pred_auto_Mod}
Let us consider a nonlinear autoregressive process with exogenous covariate $ (Y_t, \mathcal{X}_t)_{t \in \Z} $ with values in $\R \times \R$ satisfying:
\begin{equation}\label{NARX}
Y_t = f(Y_{t-1}, \ldots, Y_{t-p}; \mathcal{X}_{t-1}) + \epsilon_t,
\end{equation}
for some measurable function $f:\R^{p+1} \rightarrow \R$ ($p \in \N$), 
where $(\epsilon_t)_{t \in \Z}$ is i.i.d. generated from a standardized  uniform distribution $ \mathcal{U} [-2; 2]$ and $(\mathcal{X}_{t})_{t \in \Z}$ is an AR ($1$) process generated with a standardized $\mathcal{U} [-2; 2]$ innovations.
The process $ (Y_t, \mathcal{X}_t)_{t \in \Z} $ in (\ref{NARX}) is a specific example of the affine causal models with exogenous covariates studied in \cite{diop2022inference}. 
In the sequel,  we set $X_t =  (Y_ {t - 1}, \cdots, Y_ {t-p}, \mathcal{X}_ {t-1}) $. So, one can see that, this nonlinear autoregressive process  is a particular case of the model (\ref{regression_model}).
If $f \in \Lambda_{1}(\R^{p+1}) $ (that is, $f$ is Lipschitz) and under some classical conditions on the Lipschitz-type coefficients of $f$ (for example, $Lip(f)<1/p$), then, there exists a solution $ (Y_t, X_t)_{t \in \Z} $ from (\ref{NARX}), that fulfills the assumption \textbf{(A3)} above; see details in \cite{diop2022statistical}. 
 
\medskip
 
Let $ (Y_1, \mathcal{X}_1), \cdots, (Y_n, \mathcal{X}_n) $ be a trajectory of the process $ (Y_t, \mathcal{X}_t)_{t\in \Z}$.
 We aim to predict $Y_{n+1}$ from this training sample. 
We perform the learning theory with SPDNN predictors developed above with, the input variable $X_t =  (Y_{t - 1}, \mathcal{X}_{t-1})$, the input space $\mathcal{X}\subset \R^p \times \R $ and, the output space $\mathcal{Y} \subset \R$.
 We consider the following cases in (\ref{NARX}):
\[
\begin{array}{ll}
\text{DGP1}: & Y_t = 1 - 0.2 Y_{t-1} + 0.3 Y_{t-2} + 0.25 Y_{t-3} - 0.6 \dfrac{1}{1+\mathcal{X}_{t-1}^2} + \epsilon_t; \\
\text{DGP2}: & Y_t  = 0.5 + \big(-0.4 + 0.25e^{-2 Y_{t-1}^2}  \big) Y_{t-1}  +1.5 \mathcal{X}_{t-1} + \epsilon_t.
\end{array} 
\] 
DGP1 is a classical autoregressive model with nonlinear covariate, whereas DGP2 is an exponential autoregression with covariate.

\medskip

For each of these DGPs, we perform a network architecture of 2 hidden layers with 100 hidden nodes for each layer. The ReLU and linear activation functions are used respectively in the hidden layers and the  output layer. 
The network weights are trained in the R software with the package Keras, by using the algorithm Adam (\cite{kingma2014adam}) with learning rate $10^{-3}$ and the minibatch size of 32.
The training is stopped when the mean square error (MSE) is not improved within 30 epochs.
 %

\medskip

For $n=250, 500$ and 1000, a trajectory $ ((Y_1, \mathcal{X}_1), (Y_2, \mathcal{X}_2), \ldots, (Y_n, \mathcal{X}_n)) $ is generated from the true DGP.
The predictor $\widehat{h}_n$ is obtained from (\ref{EMR_algo}), with the tuning parameters of the form $\lambda_n = 10^{-i} \log(n)/n$ and $\tau_n = 10^{-j} / \log(n) $, where $i,j = 0, 1, \ldots, 10$ are calibrated by minimizing the MSE on a validation data set $ ((Y_1', \mathcal{X}_1'), (Y_2', \mathcal{X}_2'), \ldots, (Y_n', \mathcal{X}_n')) $.
The empirical $L_2$ error of $\widehat{h}_n$ is then computed based on a new trajectory $ ((Y_1'', \mathcal{X}_1''), (Y_2'', \mathcal{X}_2''), \ldots, (Y_m'', \mathcal{X}_m'')) $ with $m=10^4$.

\medskip

\begin{figure}[h!]
\begin{center}
\includegraphics[height=17.6cm, width=16.98cm]{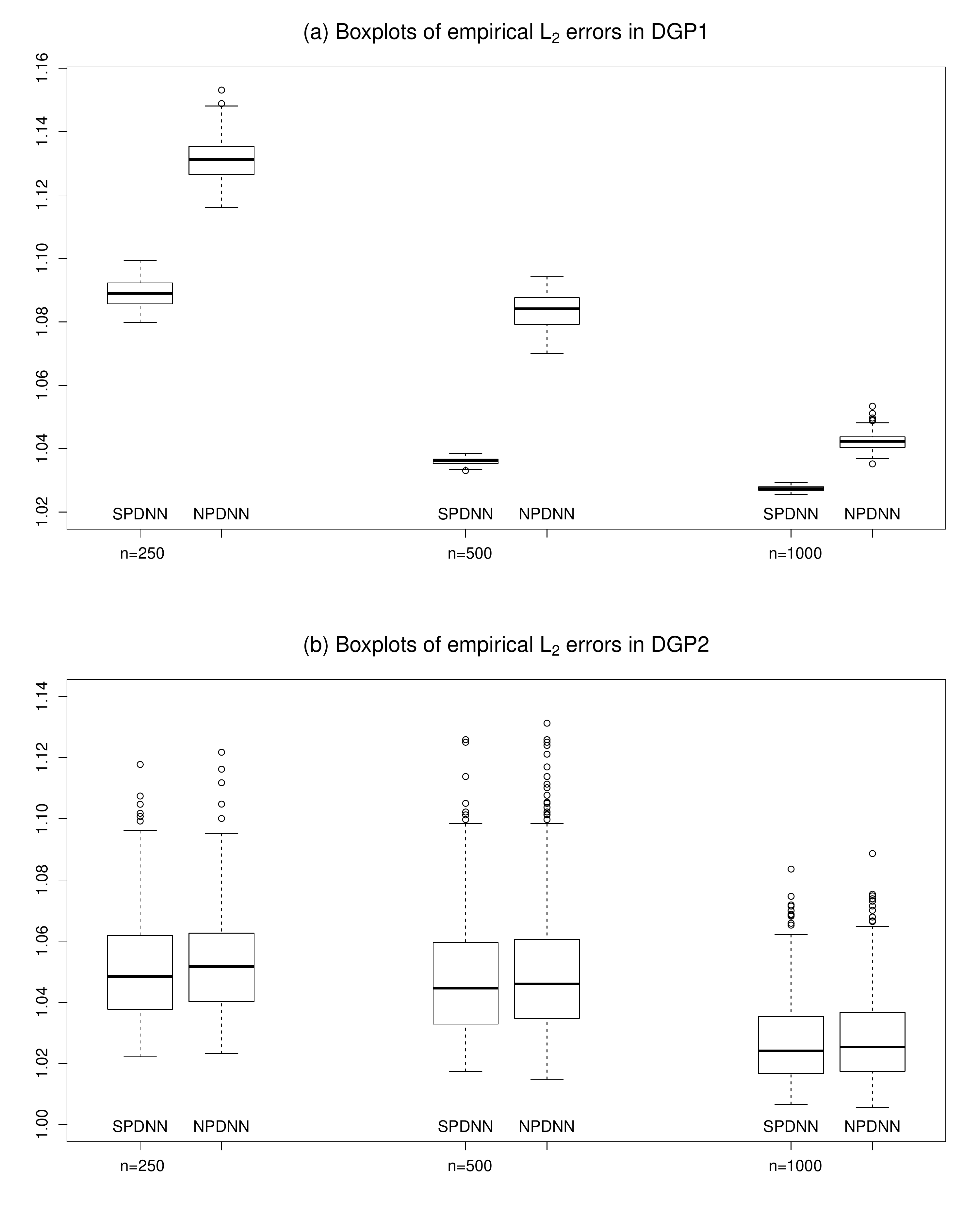}
\end{center}
\vspace{-.7cm}
\caption{\it Boxplots of empirical $L_2$ errors of the SPDNN and NPDNN predictors with $n=250, 500$ and 1000 in DGP1 (a) and DGP2 (b).}
\label{Graphe_DGP_1_2}
\end{figure}

Figure \ref{Graphe_DGP_1_2} displays the boxplots of this empirical $L_2$ error of the SPDNN and non penalized DNN (NPDNN, obtained from (\ref{EMR_algo}) with $\lambda_n=0$) predictors over 100 replications.
We  can see that, the SPDNN estimator outperforms the NPDNN estimator in DGP1.
In DGP2, the performance of the SPDNN estimator is slightly better than that of the NPDNN estimator.
These findings show that, the SPDNN estimator can improve the $L_2$ error, compared to the NPDNN estimator.

\medskip

\subsection{Prediction of binary time series}
Let us consider a binary autoregressive process with exogenous covariates $(Y_t, \mathcal{X}_t)_{t\in \Z} $ with values  in  $\{-1, 1\} \times \R$ satisfying, 
\begin{equation}\label{binary_pred}
Y_t \vert \mathcal{F}_{t - 1} \sim  2\mathcal{B} (p_t) - 1 ~ \text{with} ~ 2 p_t - 1  = \E [Y_t \vert \mathcal{F}_ {t -1 }] = f(Y_ {t-1},  \cdots,  Y_ {t - p}, ; \mathcal{X}_{t - 1}),
\end{equation}
where $\mathcal{F}_ {t-1} = \sigma \{Y_ {t-1}, \cdots ; \mathcal{X}_ {t-1}, \cdots\}$ for some measurable function $f: \R^{p+1} \rightarrow [-1, 1]$ $(p \in \N)$, $(\mathcal{X}_t)_{t\in \Z}$ is an AR(1) process and $\mathcal{B} (p_t)$ denotes the Bernoulli distribution with parameter $p_t$.
Set $X_t =  (Y_ {t - 1}, \cdots, Y_ {t-p}, \mathcal{X}_ {t-1}) $. Therefore, one can see that, the binary model (\ref{binary_pred}) is a specific case of (\ref{classifier_model}). Under some classical Lipschitz-type condition on $f$, the process $(Y_t, \mathcal{X}_t)_{t\in \Z} $ fulfills the weak dependence  assumption \textbf{(A3)}, see \cite{kengne2023deep}.

\medskip

Let $(Y_1, \mathcal{X}_1), \cdots, (Y_n, \mathcal{X}_n)$ be a trajectory of the process $(Y_t, \mathcal{X}_t)_ {t \in \Z} $, and the aim is to predict $ Y_{n+1} $ from this training sample.
We perform the learning theory with SPDNN predictor proposed here with
$p = 1, ~ X_t = (Y_ {t - 1}, \mathcal{X}_{t - 1})$ in the DGP3, and $p = 2,  ~ X_t = (Y_{t-1}, Y_{t-2}, \mathcal{X}_ {t-1})$ in the DGP4. We study the following cases of (\ref{binary_pred}):
\[
\begin{array}{ll}
\text{DGP3}:    &  f(Y_ {t-1}; \mathcal{X}_ {t-1})  = -0.15 + \big( 0.1 -0.2e^{-0.5Y_{t-1}^2}  \big) Y_{t-1} + 0.25 \dfrac{1}{ 1 + \mathcal{X}_{t - 1}^2} \\
\text{DGP4}:    & f(Y_ {t-1}, Y_ {t-2} ; \mathcal{X}_ {t-1}) = 0.1 + 0.15 Y_ {t - 1} -0.25 Y_ {t - 2}  -0.2e^{-\mathcal{X}_ {t-1}^2}.
\end{array}
\]
 In the following, we consider the  hinge loss  function $(\ell(z) = \max(1-z, 0))$  and defined 
\begin{equation}
h_{\ell}^{*} (X_t) = 2 \ind_{\{f(X_t) \ge 0\}} - 1 ~ \text{for all} ~ t\in \Z.
\end{equation}
One can see that, $h_{\ell}^{*}$ is the Bayes classifier with respect to this loss; that is                
\[ \mathcal{E}_{Z_0} (h_{\ell}^{*})  = \underset{h\in \mathcal{F}(\mathcal{X}, \mathcal{Y})}{\inf} \mathcal{E}_{Z_0} (h),  \]
where $\mathcal{E}_{Z_0}(\cdot)$ is defined in (\ref{def_excess_risk}) 
and $ \mathcal{F}(\mathcal{X}, \mathcal{Y}) $ is the class of measurable functions from $\mathcal{X} ~ \text{to} ~ \mathcal{Y} $.
Also, for each of these DGPs, we perform a network architecture as in Subsection (\ref{Pred_auto_Mod}).

\medskip

For $n=250, 500 ~ \text{and}~ 1000$, a trajectory $((Y_1, \mathcal{X}_1), \ldots, (Y_n, \mathcal{X}_n))$  is generated from the true DGP. The predictor $\widehat{h}_n$ is obtained from (\ref{def_SPDNN_class}), where the tuning parameters $\lambda_n, ~ \tau_n$ are chosen as in Subsection (\ref{Pred_auto_Mod}), based on a validation data set $ ((Y_1', \mathcal{X}_1'), (Y_2', \mathcal{X}_2'), \ldots, (Y_n', \mathcal{X}_n')) $.
The empirical excess risk of $\widehat{h}_n$ is computed from a new trajectory $ ((Y_1'', \mathcal{X}_1''), (Y_2'', \mathcal{X}_2''), \ldots, (Y_m'', \mathcal{X}_m'')) $ with $m=10^4$.

\medskip

\begin{figure}[h!]
\begin{center}
\includegraphics[height=16.8cm, width=16.98cm]{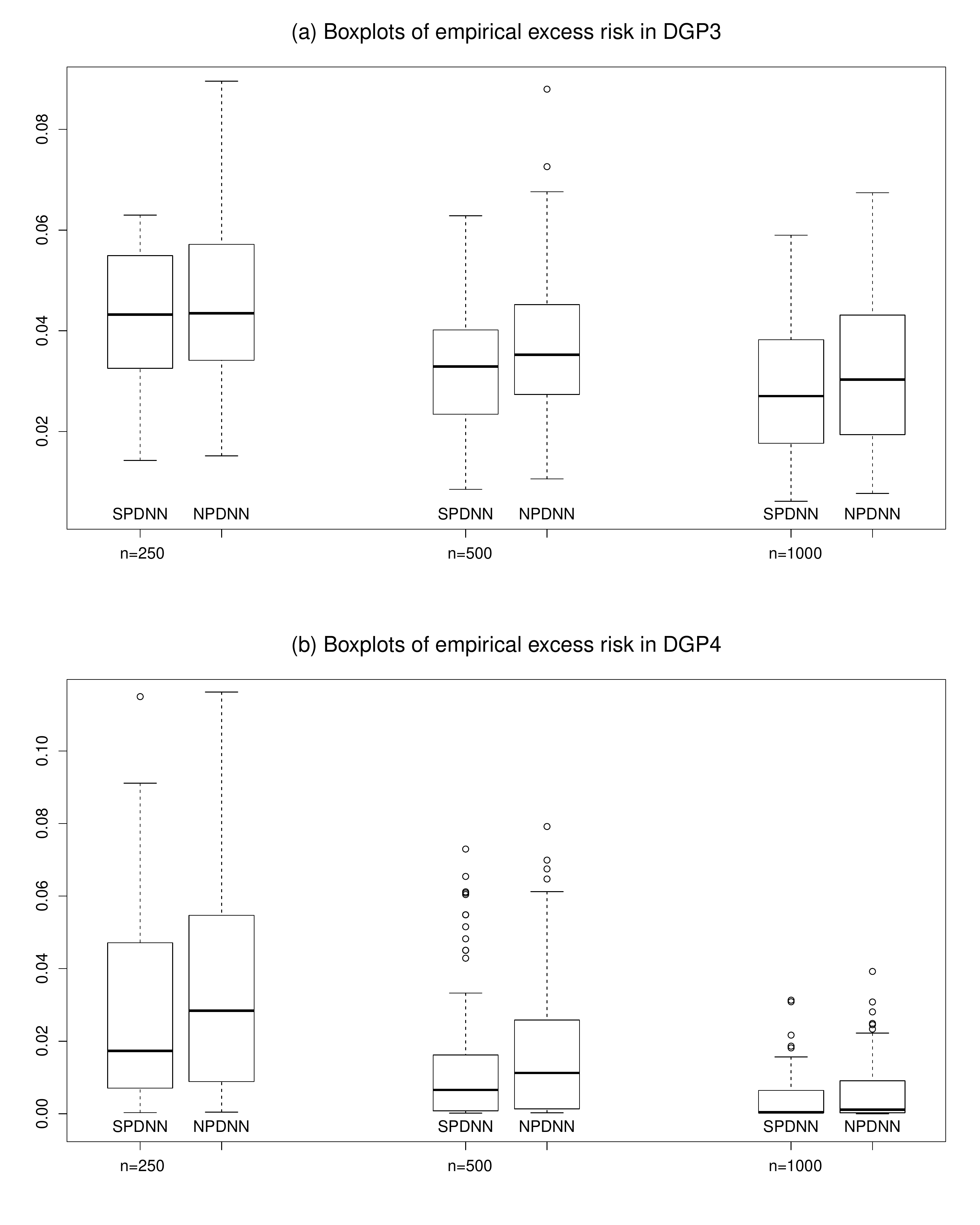}
\end{center}
\vspace{-.7cm}
\caption{\it Boxplots of the empirical  excess risk  of the SPDNN and NPDNN predictors with $n=250, 500$ and 1000 in DGP3 (a) and DGP4 (b).}
\label{Graph_DGP3_DGP4}
\end{figure}

Figure \ref{Graph_DGP3_DGP4} displays  the boxplots of the empirical excess risk of the SPDNN predictor and that of the NPDNN predictor overs 100 replications.
One can observe that, the performance of the SPDNN overall better than that of the NPDNN. Which shows once again that, the SPDNN estimator can improve the prediction accuracy, compared to the NPDNN estimator.

\section{Proofs of the main results}\label{prove}
\subsection{Proof of Theorem \ref{thm1}}
 For the proof, we  write $ a_n \lesssim_{\rho, H^ {*}} b_n$  if there is a constant $C_{\rho, H^ {*}} > 0$  depending only on $\rho$ and $H^ {*} $  such that $a_n \leq  C_{\rho, H^ {*}} b_n$ for any $n \in \N$.
Let $K_n \coloneqq (\sqrt{32 \rho^2} (\log n)^ {1/2} )  \lor  H^ {*} $. 
 Let $Y_0^ {\bot} \coloneqq \text{sign}(Y_0) (|Y_0| \land  K_n) $,  which  is  a truncated version of  $Y_0$ and  $h^{\bot} $ be the regression function of $Y_0^{\bot} $,  that is, 
 \[ h^ {\bot}(x) \coloneqq \E (Y_0^ {\bot} \vert X_0 =x).  \]
 For national convenience, we  suppress the dependency on $n$ in the  notation $Y_0^{\bot} $ and $h^{\bot} $. 
 As in \cite{ohn2022nonconvex}, consider the following decomposition 
 \begin{equation}\label{equ_risk}
\| \widehat{h}_n - h^ {*} \|_{2, P_ {X_0}}^2 = \E  \Big[ \left(Y_0 - \widehat{h}_n (X_0) \right)^2  \Big]  -  \E \Big[ \left(Y_0  - h^ {*}(X_0) \right)^2  \Big] = \sum_{i=1}^4 A_{i, n},
 \end{equation}
 where, 
 \begin{align*}
A_ {1, n} & \coloneqq \Big[ \E \left(Y_0 - \widehat{h}_ {n} (X_0)\right)^2   -  \E \left( Y_0  - h^ {*} (X_0) \right)^2  \Big] - \Big[ \E \left( Y_0^ {\bot} - \widehat{h}_ {n} (X_0) \right)^2  -  \E \left(Y_0^ {\bot}  - h^ {\bot} (X_0) \right)^2 \Big]; 
\\
A_ {2, n} & \coloneqq \Big[\E \left(Y_0^ {\bot} - \widehat{h}_ {n} (X_0) \right)^2  -  \E \left(Y_0^ {\bot}  - h^ {\bot} (X_0) \right)^2 \Big] 
- 2 \Big[ \frac{1}{n} \sum_{i=1}^n \left( Y^ {\bot}_i - \widehat{h}_n (X_i) \right)^2  -  \frac{1}{n} \sum_{i=1}^n \left(Y^ {\bot}_i  - h^ {\bot} (X_i) \right)^2 \Big] - 2 J_{\lambda_n,  \tau_n} (\widehat{h}_n);  
\\
A_ {3, n}  & \coloneqq 2 \Big[ \frac{1}{n} \sum_{i=1}^n \left(Y^ {\bot}_i - \widehat{h}_n (X_i) \right)^2 -  \frac{1}{n} \sum_{i=1}^n \left(Y^ {\bot}_i - h^ {\bot}(X_i) \right)^2 \Big]   
- 2 \Big[ \frac{1}{n} \sum_{i=1}^n \left(Y_i - \widehat{h}_{n} (X_i) \right)^2 -  \frac{1}{n} \sum_{i=1}^n \left(Y_{i} - h^ {*}(X_i) \right)^2 \Big]; 
\\
A_ {4, n} & \coloneqq 2 \left[ \frac{1}{n} \sum_{i=1}^n \left(Y_i - \widehat{h}_n (X_i) \right)^2 - \frac{1}{n} \sum_{i=1}^n \left(Y_i - h^ {*} (X_i) \right)^2  \right] + 2J_{\lambda_n, \tau_n}(\widehat{h}_n). 
\end{align*}
The first equality in (\ref{equ_risk}) holds by using the independence of $X_0$ and $\epsilon_0$.
To bound $A_ {1, n} $,  let us recall the properties of sub-Gaussian  variables  $ \E (\epsilon_0) = 0$ and $\E (e^ {\epsilon_0^2/4\rho^2}) \leq \sqrt{2} $, see, for instance, Theorem 2.6 in \cite{wainwright2019high}. 
Let 
\begin{align*}
A_ {1, 1, n} & \coloneqq \E \left((Y_0^ {\bot} - Y_0)(2\widehat{h}_n (X_0) - Y_0 - Y_0^ {\bot}) \right);  
\\
A_ {1, 2, n}  & \coloneqq \E \left[(Y_0^ {\bot} -h^ {\bot} (X_0) - Y_0 + h^ {*}(X_0) )(Y_0^ {\bot} - h^ {\bot} (X_0) + Y_0 - h ^{*} (X_0) ) \right]. 
\end{align*}
One can see that $A_ {1, n} = A_ {1, 1, n} + A_ {1, 2, n} $.  We use the Cauchy-Schwarz inequality to obtain 
\[ |A_ {1, 1, n}| \leq \sqrt{E(Y_0^ {\bot} - Y_0)^2} \sqrt{ \E (2 \widehat{h}_n (X_0) - Y_0 - Y_0^ {\bot})^2}.   \]
We have,
\begin{align*}
(Y_0^ {\bot} - Y_0)^2 & = (|Y_0| \land K_ {n})^2 - 2 |Y_0| (|Y_0| \land K_ {n}) + |Y_0|^2
\\
& = (|Y_0|  - K_ {n})^2 \ind{(|Y_0| > K_ {n})}.
\end{align*}
Hence, 
\begin{align}\label{equ_Bd}
\E (Y_0^{\bot} - Y_0)^2  \nonumber & = \E \left[ (|Y_0| - K_n)^2 \ind{(|Y_0| > K_n)} \right]
\\
 & \leq  \E \left[ |Y_0|^2 \ind{(|Y_0|> K_n)}  \right].  
 \end{align}
 From the assumption (\ref{def_H_rho_H_star}) and the independence of $X_0$ and $\epsilon_0$, we get,
 $ \E (e^ {Y_0^2/(8 \rho^2)}) \leq e^ {(H^ {*}) ^2/(4 \rho^2)} \E e^{\epsilon_0^2 / (4\rho)} \leq \sqrt{2} e^ {(H^ {*}) ^2/(4 \rho^2)} $.
 Also, one can easily see that, $ Y_0^2 \leq 16 \rho^2 e^ {Y_0^2 /16\rho^2} ~ \text{and} ~ \ind{(|Y_0| > K_n)} \leq  e^ {(Y_0^2 - K_n) /16 \rho^2} $.
 Therefore,
 \begin{align}
\E(Y_0^ {\bot} - Y_0) ^2  \nonumber & \leq \E \left[16 \rho^2 e^ {Y_0^2 /(16 \rho^2)} e^ {Y_0^2/(16 \rho^2) - K_n^2 /(16 \rho^2)} \right] 
\\
 &  \leq 16 \sqrt{2} \rho^2 e^ {(H^ {*}) ^2 /(4 \rho^2)} e^ {-2 \log n} \leq 16 \sqrt{2} \rho^2 e^ {(H^ {*}) ^2 /(4 \rho^2)} n^ {-2}. 
\end{align}
We also have, 
\begin{align}\label{equ_Bd2}
\E(2 \widehat{h}_n (X_0) - Y_0 - Y_0^ {\bot}) ^2 \nonumber & \leq 2 \E(Y_0^2) + 2 \E (2 \widehat{h}_n (X_0) - Y_0^ {\bot}) ^2,
\end{align}
and, 
\begin{align*}
|(2 \widehat{h}_n (X_0) - Y_0^ {\bot}) ^2 | & \leq 4 | \widehat{h}_n (X_0)|^2 + 4 |\widehat{h}_n (X_0) | |Y_0^{\bot}| + |Y_0^{\bot}|^{2} 
\\
& \leq 4 K_n^2 + 4 K_n^2 + K_n^2.
\end{align*}
Since, there exists a constant $C$ such that $K_n^2 \leq  C \log n$,  we have 
\begin{align}
\E( 2 \widehat{h}_n (X_0) - Y_0 - Y_0^{\bot}) ^2 
\nonumber   & \leq 16 \rho^2 \E (e^ {Y^2 /(8 \rho^2)}) + 18K_n^2 
\\
&  \lesssim C_{\rho, H^ {*}} \log n. 
\end{align}
Hence, $|A_ {1, 1, n}| \lesssim C_{\rho, H^ {*}} \log n/n$.  
Let us deal now with $A_ {1, 2, n} $. 
By using the Cauchy-Schwarz inequality, we get, 
\begin{align*}
|A_ {1, 2, n}| \leq \sqrt{ 2 \E (Y_0^{\bot} - Y_0) ^2 + 2 \E(h^{\bot} (X_0) -  h^ {*} (X_0)) ^2} \times \sqrt{ \E (Y_0 + Y_0^{\bot} - h^{\bot} (X_0) - h^ {*}(X_0)) ^2}.
\end{align*}
In a similar way as in (\ref{equ_Bd2}) we get, $ \E (Y_0 + Y_0^ {\bot} - h^ {\bot} (X_0) - h^ {*} (X_0)) ^2 \lesssim_{C_{\rho}, H^ {*}} \log n$. 
Since $ \E (\epsilon_0) = 0$, from the Jensen' s inequality, one can easily get, 
\[ \E \big[(h^{\bot} (X_0) - h^ {*} (X_0)) ^2 \big]  = \E \left( \E (Y_0^{\bot} \vert X_0) - \E (Y_0 \vert X_0) \right) ^2 \leq  \E (Y_0^ {\bot} - Y_0) ^2. \]
By using (\ref{equ_Bd}), one can obtain $ |A_ {1, 2, n} | \lesssim C_{\rho, H^ {*}} \log n/n$.  
Now, we have, 
\[ \E[A_ {3, n}] = -\dfrac{2}{n} \sum_{i=1}^n \left\{ \left[ \E[(Y_i - \widehat{h}_n (X_i) ) ^2] - \E[(Y_i - h^ {*} (X_i) ) ^2]  \right]  
- \left[\E[(Y_i^ {\bot} - \widehat{h}_{n} (X_i) ) ^2] - \E[(Y_i^ {\bot} - h^ {\bot} (X_i) ) ^2]  \right] \right\}.  \]
For $i= 1, \cdots, n$, set, 
\begin{align*}
 A_ {3, n, i}&=  \left[ \E [(Y_i - \widehat{h}_{n} (X_i) ) ^2]- \E[ (Y_i - h^ {*} (X_i)) ^2] \right] - \left[\E[(Y_i^ {\bot} - \widehat{h}_{n} (X_i) ) ^2] - \E[(Y_i^ {\bot} - h^ {\bot} (X_i)) ^2] \right] \\
A_ {3, 1, n, i} & = \E \left[ (Y_i^ {\bot} - Y_i)(2 \widehat{h}_n (X_i) - Y_i - Y_i^ {\bot})  \right] 
\\
A_ {3, 2, n, i} & = \E\left[ (Y_i^ {\bot} - h^ {\bot} (X_i) - Y_i + h^ {*}(X_i) )(Y_i^ {\bot} - h^ {\bot} (X_i)  + Y_i - h^ {*} (X_i))  \right]. 
\end{align*}
We have, for $ i=1, \cdots, n$,  
\[ | A_ {3, 1, n, i} | \leq  \sqrt{\E \left[ (Y_i^ {\bot}  - Y_i) ^2 \right]} \sqrt{\E \left[(2 \widehat{h}_n (X_i) - Y_i  - Y_i^ {\bot}) ^2 \right]}. \]
By using similar arguments as in $A_ {1, 1, n} $, we get for $i=1, \cdots, n$,   
\[ | A_ {3, 1, n, i}| \lesssim C_{\rho, H^ {*}} \log n/n.  \] 
Also, by going as in $A_ {1, 2, n} $, it holds for $i=1, \cdots, n$, that,  
\[ |A_ {3, 2, n, i}| \leq \sqrt{\E \left[ (Y_i^ {\bot} - h^ {\bot} (X_i) - Y_i + h^ {*} (X_i)) ^2 \right]} \sqrt{\E \left[(Y_i^ {\bot} - h^ {\bot} (X_i) + Y_i  - h^ {*} (X_i) )^2 \right]},  \]
and we can also obtain $|A_ {3, 2, n, i} | \lesssim C_{\rho, H^ {*}} \log n/n$. 
Thus, for $i =1, \cdots, n$,  
\[ |A_ {3, n, i} |  \lesssim C_{\rho, H^ {*}} \log n/n.  \] 
Hence, 
\begin{align}
|\E (A_ {3, n})|  \nonumber  = |-\dfrac{2}{n} \sum_{i=1}^n A_ {3, n, i} |
 \leq \dfrac{2}{n} \sum_{i=1}^n |A_ {3, n, i} | 
  \lesssim C_{\rho, H^ {*}} \log n/n.
\end{align}
%

For $A_ {2, n} $,  define  $\Delta h (Z_0) \coloneqq  (Y_0^ {\bot}  - h (X_0) ) ^2 - (Y_0^ {\bot}  - h^ {\bot} (X_0) ) ^2 $  with  $ Z_0 \coloneqq (X_0, Y_0) $ for  $h \in \mathcal{H}_ {\sigma}(L_n, N_n, B_n, F) $.
Let $\alpha > 0$,  we can write  
\begin{align*}
P(A_ {2, n} > \alpha) & \leq P\left(\underset{h \in \mathcal{H}_ {\sigma}(L_n, N_n, B_n, F)}{\sup} \frac{ \E(\Delta h (Z_0)) - \frac{1}{n}\sum_{i=1}^n \Delta h (Z_i)}{\alpha + 2J_{\lambda_n, \tau_n} + \E(\Delta h (Z_0))} \ge \frac{1}{2}\right) 
\\
 & \leq \sum_{j=0}^ {\infty} P\left(\underset{h \in \mathcal{H}_ {n, j, \alpha}}{\sup} \dfrac{ \E (\Delta h (Z_0)) - \dfrac{1}{n} \sum_{1=1}^n \Delta h (Z_i) }{ 2^j \alpha + \E (\Delta h (Z_0) )} \ge \dfrac{1}{2} \right),
\end{align*} 
where 
\begin{equation}
\mathcal{H}_ {n, j, \alpha} \coloneqq  \Big\{h \in \mathcal{H}_ {\sigma}(L_n, N_n, B_n, F): 2^{j - 1} \ind{(j \ne 0)} \alpha \leq  J_{\lambda_n, \tau_n}(h) \leq 2^j \alpha  \Big\}. 
\end{equation}

Indeed,
\begin{align*}
\nonumber & P (A_ {2, n} > \alpha)  
\\
\nonumber & = P\Bigg(\left[\E \left(Y_0^ {\bot} - \widehat{h}_n (X_0) \right) ^2   -  \E \left(Y_0^ {\bot}  - h^ {\bot} (X_0) \right) ^2 \right] 
- 2 \left[ \dfrac{1}{n} \sum_{i=1}^n \left(Y_i^ {\bot} - \widehat{h}_n (X_i) \right) ^2  -  \dfrac{1}{n} \sum_{i=1}^n \left(Y_i^ {\bot}   - h^ {\bot} (X_i) \right) ^2 \right] \\
  & \hspace{14cm} - 2  J_{\lambda_n, \tau_n}(\widehat{h}_n) > \alpha \Bigg) 
\\
\nonumber & \leq P\Bigg(2 \left[\E \left(Y_0^ {\bot} - \widehat{h}_n (X_0) \right) ^2  -  \E \left(Y_0^ {\bot}  - h^ {\bot} (X_0) \right) ^2 \right] 
 - \left[\dfrac{1}{n} \sum_{i=1}^n \left(Y_i^ {\bot}  - \widehat{h}_n (X_i) \right) ^2   -  \dfrac{1}{n} \sum_{i=1}^n \left(Y_i^ {\bot}  - h^{\bot} (X_i) \right) ^2 \right] 
 \\
\nonumber & \hspace{7.4cm} > \alpha + 2 J_{\lambda_n, \tau_n}(\widehat{h}_n) + \E \left(Y_0^ {\bot}  - \widehat{h}_n (X_0) \right) ^2 -  \E\left(Y_0^ {\bot}  - h^ {\bot} (X_0) \right) ^2 \Bigg) 
\\
\nonumber  & \leq  P\Bigg( \exists h \in \mathcal{H}_ {\sigma}(L_n, N_n, B_n, F) : \left[ \E \left(Y_0^ {\bot} - h(X_0) \right) ^2  -  \E \left(Y_0^ {\bot}  - h^ {\bot} (X_0) \right) ^2 \right]  \\
& \hspace{9.2cm} - \left[\dfrac{1}{n} \sum_{i=1}^n \left(Y_i^{\bot} - h (X_i)  \right) ^2  -  \dfrac{1}{n} \sum_{i=1}^n \left(Y_i^ {\bot}  - h^{\bot} (X_i) \right) ^2 \right]  
 \\
\nonumber  & \hspace{6.2cm} > \dfrac{1}{2} \left(\alpha + 2 J_{\lambda_n, \tau_n}(h (X_0)) + \E \left(Y_0^ {\bot} - h (X_0) \right) ^2  -  \E \left(Y_0^ {\bot}  - h^ {\bot} (X_0) \right) ^2 \right) \Bigg)
\\
 \nonumber  & \leq  P\Bigg( \exists h \in \mathcal{H}_ {\sigma}(L_n, N_n, B_n, F): \\
 \nonumber  & \hspace{1.5cm}  \dfrac{\left[ \E \left(Y_0^ {\bot} - h (X_0) \right) ^2   -  \E \left(Y_0^ {\bot}  - h^ {\bot} (X_0) \right) ^2  \right] 
 - \left[ \dfrac{1}{n} \sum_{i=1}^n \left(Y_i^{\bot}  - h (X_i) \right) ^2  -  \dfrac{1}{n} \sum_{i=1}^n \left(Y_i^{\bot}  - h^{\bot} (X_i) \right) ^2 \right] }{\left( \alpha + 2 J_{\lambda_n, \tau_n} (h (X_0)) + \E\left(Y_0^{\bot} - h (X_0) \right) ^2 -  \E \left(Y_0^{\bot}  - h^ {\bot} (X_0) \right) ^2 \right)}  \ge \dfrac{1}{2}  \Bigg) 
 \\
\nonumber  & \leq \sum_{j=0}^{\infty} P\Bigg( \underset{h \in \mathcal{H}_ {\sigma}(L_n, N_n, B_n, F)}{\sup} \\ 
& \hspace{1.5cm}  \dfrac{\left[ \E\left(Y_0^ {\bot} - h (X_0) \right) ^2  -  \E \left(Y_0^ {\bot}   - h^ {\bot} (X_0) \right) ^2 \right] 
 - \left[\dfrac{1}{n} \sum_{i=1}^n \left(Y_i^{\bot}  - h (X_i) \right) ^2  -  \dfrac{1}{n} \sum_{i=1}^n \left(Y_i^ {\bot} - h^ {\bot} (X_i) \right) ^2 \right] }{\left(2^j \alpha + 2 J_{\lambda_n, \tau_n}(h (X_0)) + \E \left(Y_0^{\bot} - h (X_0) \right) ^2 -  \E \left(Y_0^{\bot}  - h^ {\bot} (X_0) \right) ^2 \right)}  \ge \dfrac{1}{2}  \Bigg) 
  \\
  & \leq  \sum_{j =0}^{\infty} P\Bigg( \underset{h \in\mathcal{H}_ {n, j, \alpha}}{\sup} \dfrac{\E \Delta h (X_0) - \dfrac{1}{n} \sum_{i=1}^n \Delta h (X_i) }{\left(2^j\alpha + 2 J_{\lambda_n, \tau_n}(h (X_0)) + \E \Delta h (X_0) \right)} 
  \ge \dfrac{1}{2} \Bigg).
 \end{align*} 

 \medskip

 \noindent For $h \in \mathcal{H}_ {n, j, \alpha}$,  we have 
\[ P\Bigg( \underset{h \in \mathcal{H}_ {\sigma}(L_n, N_n, B_n, F)}{\sup} \frac{\E \Delta h (Z_0) - \dfrac{1}{n} \sum_{i=1}^n \Delta h (Z_i)}{\left( \alpha + 2 J_{\lambda_n, \tau_n}(h (X_0)) + \E \Delta h (Z_0) \right)} 
\ge \dfrac{1}{2} \Bigg) \leq \sum_{j=0}^ {\infty} P\Bigg( \underset{h \in \mathcal{H}_ {n, j, \alpha}}{\sup} \dfrac{\E \Delta h (Z_0) - \dfrac{1}{n} \sum_{i=1}^n \Delta h (Z_i)}{\left( 2^j \alpha  + \E  \Delta h (Z_0) \right)} \ge \dfrac{1}{2}  \Bigg).  \]
One can easily show that, 
\begin{align*}
\E (\Delta h(Z_0)]   & =   \E [(h (X_0) - h^ {\bot} (X_0) ) ^2] -\E [(h^ {\bot} (X_0)) ^2] +  2 \E [(h^ {\bot} (X_0)) ^2] - \E[(h^ {\bot}) ^2 (X_0)]
\\
  &=  \E[(h(X_0) - h^ {\bot} (X_0) )^{2}] \ge 0.
\end{align*}
Hence, 
\begin{align}\label{equ_D}
\nonumber & \sum_{j=1}^\infty  P\Bigg( \underset{h \in \mathcal{H}_ {n, j, \alpha}}{\sup} \frac{\E \Delta h (Z_0) - \dfrac{1}{n}\sum_{i=1}^n \Delta h (Z_i)}{\left( 2^j \alpha + \E \Delta h (Z_0) \right)} \ge \dfrac{1}{2} \Bigg)
 \leq  \sum_{j=1}^\infty P\Bigg( \underset{h \in \mathcal{H}_ {n, j, \alpha}}{\sup} \frac{\E \Delta h (Z_0) - \dfrac{1}{n} \sum_{i=1}^n \Delta h (Z_i)}{ 2^j \alpha } \ge \dfrac{1}{2}  \Bigg)
\\
& \nonumber  \leq  \sum_{j=1}^\infty  P\Bigg( \underset{h \in \mathcal{H}_ {n, j, \alpha}}{\sup} \left\{\E \Delta h (Z_0) - \dfrac{1}{n} \sum_{i=1}^n \Delta h(Z_i) \right\} \ge \dfrac{2^j \alpha}{2}  \Bigg)
 \leq  \sum_{j=1}^\infty P\Bigg( \underset{g \in \mathcal{G}_ {n, j, \alpha}}{\sup} \left\{\E[g (Z_0)]- \frac{1}{n} \sum_{i=1}^n g  \right\} \ge \frac{2^j \alpha}{2}  \Bigg),
\end{align}
where, 
\begin{equation}
\mathcal{G}_ {n, j, \alpha} \coloneqq \Big\{ \Delta (h): \R^{d} \times\R \rightarrow \R, h \in \mathcal{H}_ {n, j, \alpha} \Big\}. 
\end{equation}
Let $h \in \mathcal{H}_ {\sigma}(L_n, N_n, B_n, F)$. Consider the function $g(x,y) = \Delta h(x,y)$. One can easily prove that, $g$ is Lipschitz with Lipschitz coefficient $\max\big(2F Lip(h) + 2K_n Lip(h), 6 K_n + 2F \big)$.  
Therefore, one can get that, the process $ (g (Z_t))_{t \in \Z} $ is also $\psi$-weakly dependent. 
Thus, we have from \cite{hwang2014note} (see also \cite{diop2022statistical}),
\begin{align}\label{equ_b}
 P\left\{\E[g (Z_0)] - \dfrac{1}{n}\sum_{i=1}^n g (Z_i) > \varepsilon \right\} \nonumber & =  P\left\{ \sum_{i=1}^n \left(\E[g (X_0, Y_0)] -  g (X_i, Y_i) \right) \ge  
 n \varepsilon  \right\} 
 \\
 \nonumber  & \leq P\left\{ \left|\sum_{i=1}^n \left(\E [g (X_0, Y_0)] - g (X_i, Y_i)  \right) \right| \ge  
 n \varepsilon \right\} 
 \\
 \nonumber & \leq C_3 \log n \exp\left(-\dfrac{n^2 \varepsilon^2}{A'_n + B'_n (n \varepsilon)^ {\nu}} \right) 
 \\
 & \leq C_{3} \exp\left(\log \log n -\dfrac{n^2 \varepsilon^2}{A'_n + B'_n (n \varepsilon)^ {\nu}} \right).
\end{align}
For a some  constant $C_3 > 0$, any sequence $ (A'_n)_ {n \in \N} $, satisfying   $A'_n \ge \E \left[\left(\sum_{i=1}^n \left(g (X_i, Y_i) - \E [g (X_0, Y_0)] \right) \right) ^2 \right]$ and $ B_n = \frac{n^ {3/4} \log n}{A'_n} $.  Let $l = \mathcal{N}( \varepsilon, \mathcal{G}_ {n, j, \alpha}, \| \cdot \|_\infty) $.  For n large enough, we have 
\begin{align}\label{equ_br}
P\left\{\underset{ g \in \mathcal{G}_{n, j, \alpha}}{\sup} \left[ \E[g(Z_0)] - \dfrac{1}{n} \sum_{i=1}^n g(Z_i) \right] > \varepsilon \right\} 
\nonumber & \leq  C_{3} \sum_{i=1}^l \exp\left(\log \log n -\dfrac{n^2 \varepsilon^2/4}{A'_n  +  B'_n (n \varepsilon/2)^\nu} \right)   
\\
\nonumber & \leq C_{3}\cdot l \exp\left(\log \log n -\frac{n^2 \varepsilon^2/4}{A'_n  +  B'_n (n \varepsilon/2)^\nu} \right)
\\
\nonumber & \leq C_{3}\mathcal{N}(\varepsilon, \mathcal{G}_{n, j, \alpha}) \exp\left(\log \log n - \frac{n^2 \varepsilon^2/4}{A'_n + B'_n (n \varepsilon/2)^\nu} \right).
\end{align}
 In \cite{ohn2022nonconvex}, we have  
\begin{equation}\label{P_inqu2}
\mathcal{N}(\varepsilon, \mathcal{G}_{n, j, \alpha}, \| \cdot \|_\infty) \leq \mathcal{N} (\frac{\varepsilon}{4 K_n}, \mathcal{H}_{n, j, \alpha}, \| \cdot \|_\infty).
\end{equation} 
One can easily see that,
\begin{equation}\label{inclusion}
 \mathcal{H}_{n, j, \alpha}  \subset \left\{  h \in \mathcal{H}_{\sigma}(L_{n}, N_{n}, B_{n}, F, \dfrac{2^j \alpha}{\lambda_n}): \| \theta(h) \|_{ \text{clip}, \tau_n} \leq \frac{2^j \alpha}{\lambda_n}  \right\}. 
 \end{equation}
Thus, in \cite{ohn2022nonconvex}, we have the following inequality
\begin{align}\label{cover_number}
\mathcal{N}(\varepsilon, \mathcal{G}_{n, j, \alpha}, \| \cdot \|_\infty)  \nonumber & \leq \mathcal{N}(\dfrac{\varepsilon}{4 K_n}, \mathcal{H}_{n, j, \alpha}, \| \cdot \|_\infty) 
\\
\nonumber & \leq \mathcal{N}(\frac{\varepsilon}{4 K_n}, \mathcal{H}_{\sigma}(L_{n},N_{n}, B_{n}, F, \frac{2^j \alpha}{\lambda_n}),  \| \cdot \|_\infty )
\\
& \leq  \exp\left(2 \frac{2^j \alpha}{\lambda_n}(L_n + 1) \log \left(\frac{(L_n + 1)(N_n + 1)B_n}{\dfrac{\varepsilon}{4K_n} - \tau_n (L_n + 1)((N_n + 1) B_n)^{L_n +1}} \right) \right).
\end{align} 
We have,
\begin{align}
\nonumber & P\Big\{\underset{ g \in \mathcal{G}_{n, j, \alpha}}{\sup} \left[\E [g(Z_0)] - \dfrac{1}{n} \sum_{i=1}^n g(Z_i) \right] > \varepsilon \Big\} 
\\
& \leq C_{3} \exp\left( 2 \frac{2^j \alpha}{\lambda_n}(L_n + 1) \log \left(\frac{(L_n + 1)(N_n + 1)B_n }{\frac{\varepsilon}{4 K_n} - \tau_n (L_n + 1)((N_n + 1) B_n)^{L_n +1}} \right) + \log \log n -\frac{n^2 \varepsilon^2/4}{A'_n + B'_n (n \varepsilon/2)^\nu} \right).
\end{align}
Hence,
\begin{align}
\nonumber  & \sum_{j=1}^\infty P\left\{\underset{ g \in \mathcal{G}_{n, j, \alpha}}{\sup} \left[\E [g(Z_0)] - \dfrac{1}{n} \sum_{i=1}^n g(Z_i) \right] > \varepsilon \right\}
\\
 & \leq C_{3}  \sum_{j=1}^\infty \exp\left( 2 \frac{2^j \alpha}{\lambda_n}(L_n + 1) \log \left(\frac{(L_n + 1)(N_n + 1)B_n }{\frac{\varepsilon}{4K_n} - \tau_n (L_n + 1)((N_n + 1) B_n)^{L_n +1}} \right) + \log \log n -\frac{n^2 \varepsilon^2/4}{A'_n + B'_n (n \varepsilon/2)^\nu} \right).
\end{align}
Let $\varepsilon = \dfrac{2^j \alpha}{2}$,  we have
\begin{align}
\nonumber & \sum_{j=1}^\infty P\Big\{\underset{ g \in \mathcal{G}_{n, j, \alpha}}{\sup} \left[\E [g(Z_0)]  - \dfrac{1}{n} \sum_{i=1}^n g(Z_i) \right] >  \frac{2^j \alpha}{2} \Big\} 
\\
\nonumber  & \leq C_{3}  \sum_{j=1}^\infty \exp\left( 2 \frac{2^j \alpha}{\lambda_n}(L_n + 1) \log\left(\dfrac{ (L_n + 1) (N_n + 1)B_n}{ \dfrac{2^j \alpha}{8 K_n} - \tau_n (L_n + 1)((N_n + 1) B_n)^{L_n +1}} \right) + \log \log n -\frac{(n2^j \alpha)^2/16}{A'_n + B'_n (n 2^j \alpha/4)^\nu} \right) 
\\
 &  \leq C_{3}  \exp( \log \log n) \sum_{j=1}^\infty \exp\left( 2 \dfrac{2^j \alpha}{\lambda_n} (L_n + 1) \log \left(\frac{ (L_n + 1)(N_n + 1) B_n}{ \dfrac{2^j \alpha}{8 K_n} - \tau_n (L_n + 1)((N_n + 1) B_n)^{L_n +1}} \right)  -\frac{(n 2^j \alpha)^2/16}{A'_n + B'_n (n 2^j \alpha/4)^\nu} \right). 
\end{align}
By with choice $A'_n = n C$ and $B'_n = \log(\frac{n}{n^{1/4}  C})$,  we have
\begin{align}\label{equ_5_18}
\nonumber  & \sum_{j=1}^\infty P\Big\{\underset{ g \in \mathcal{G}_{n, j, \alpha}}{\sup} \left[\E [g(Z_0)] - \dfrac{1}{n} \sum_{i=1}^n g(Z_i) \right] >  \frac{2^j \alpha}{2} \Big\} 
\\
\nonumber &  \leq C_{3}  \exp( \log \log n) \\
& \times \sum_{j=1}^\infty \exp\left( 2 \frac{2^j \alpha}{\lambda_n} (L_n + 1) \log \left(\frac{ (L_n + 1)(N_n + 1) B_n}{ \dfrac{2^j \alpha}{8 K_n} - \tau_n (L_n + 1)((N_n + 1) B_n)^{L_n +1}} \right)  -\frac{(n 2^j \alpha)^2/16}{n C + \log(\frac{n}{n^{1/4} C})(n 2^j \alpha/4)^\nu} \right).
\end{align}
Let, 
\begin{equation}\label{equ_step_1}
\log(\frac{n}{n^ {1/4} C})(n 2^j \alpha /8)^ {\nu} > n C \Longrightarrow \alpha > \frac{8}{n}\left(\frac{n C}{\log(\frac{n}{n^ {1/4} C})}\right)^ {1/\nu} \coloneqq \alpha_{n}.  
\end{equation}
We can easily see that from (\ref{equ_step_1}) 
\begin{equation*}
 -\frac{(n 2^j \alpha)^ {2} /16}{n C  + \log(\frac{n}{n^ {1/4} C})(n 2^j \alpha /8)^ {\nu}} \leq -\frac{(n 2^j \alpha)^ {2} /16}{2\log(\frac{n}{n^ {1/4} C})(n 2^j \alpha/4)^ {\nu}}.  
\end{equation*}
We can see also asymptotically $\alpha_{n} > 1$.

\noindent$\textbf{Step 1}: \alpha > \alpha_n$ 
Under the assumption 
\[ \tau_n \leq \dfrac{1}{16 K_n  (L_n + 1)((N_n + 1) B_n)^{L_n +1}}, \]
And for  $2^j \alpha >1$, we have
\begin{align}
 \nonumber & \sum_{j=1}^ {\infty} \exp\left( 2 \frac{2^j \alpha}{\lambda_n}(L_n + 1) \log \left( \dfrac{(L_n + 1)(N_n + 1) B_n}{\dfrac{2^j \alpha}{8 K_n} - \tau_n (L_n + 1)((N_n + 1) B_n)^{L_n +1}}  \right)  -\frac{(n 2^j \alpha)^ {2} /16}{n C + \log(\frac{n}{n^ {1/4} C})(n 2^j \alpha/4)^ {\nu}}  \right) 
 \\
  & \leq    \exp \left(2 \frac{2^j \alpha}{\lambda_n}(L_n + 1) \log \left( 16 K_n (L_n + 1)(N_n + 1) B_n  \right)  -\frac{(n 2^j \alpha)^ {2-\nu} /16}{2 \log(\frac{n}{n^ {1/4} C}) /4^ {\nu}} \right).
\end{align}
For $\alpha > \alpha_n$  we have $ 2^j \alpha  < ((2 \alpha)^{2-\nu})^j$
Thus,
\begin{align}
\nonumber & \sum_{j=1}^ {\infty} P\Big\{\underset{ g \in \mathcal{G}_ {n, j, \alpha}}{\sup} \left[\E [g (Z_0)] - \dfrac{1}{n} \sum_{i=1}^n g (Z_i) \right] >  \frac{2^j \alpha}{2} \Big\}  
\\
& \leq \sum_{j=1}^ {\infty} \exp\left( 2^j \alpha \left(\frac{2} {\lambda_n}(L_n + 1) \log \left(16 K_n (L_n + 1)(N_n + 1) B_n  \right)  -\frac{n^ {2-\nu} /32}{\log(\frac{n}{n^ {1/4} C}) /4^ {\nu}} \right) \right).
\end{align}
We can see that for n large enough,
\[  \delta_{n, 1} \coloneqq \frac{2}{\lambda_{n}}(L_{n} + 1)\log\left(16 K_{n}(L_{n} + 1)(N_{n} + 1)B_{n}  \right)  \leq \frac{n^ {2-\nu} /32}{2\log(\frac{n}{n^ {1/4} C})/4^{\nu}}. \] 
Thus,
\begin{align}
\nonumber & \sum_{j=1}^ {\infty} P\Big\{\underset{ g \in \mathcal{G}_ {n, j, \alpha}}{\sup} \left[\E [g (Z_0)] - \dfrac{1}{n} \sum_{i=1}^n g (Z_i) \right] >  \frac{2^j \alpha}{2} \Big\}  \leq C_3 \log n \sum_{j=1}^ {\infty} \exp\left( -2^j \dfrac{n^ {2-\nu}}{64 \log(\frac{n}{n^ {1/4} C})/4^ {\nu}} \alpha \right).
\end{align}
Let 
\[ \beta_n \coloneqq \dfrac{n^ {2-\nu}}{64 \log \left( \dfrac{n}{n^ {1/4} 
 C} \right)/4^ {\nu}} \alpha,   \]
 we have 
\begin{align}\label{inequ1}
   \sum_{j=0}^ {\infty} \exp\left(- \beta_n (2^j \right) &  \leq \exp\left(-\beta_n \right) \sum_{j=0}^ {\infty} \left(2^{ -\dfrac{\beta_n }{\log 2}} \right)^j  \leq \exp\left(- \beta_n \right) \left(\dfrac{1}{1 - 2^ {- \dfrac{\beta_n }{\log 2}}} \right)  \leq  \exp\left(-\beta_n \right)
\end{align}
By applying   (\ref{inequ1}),  we have 
\begin{align}
\sum_{j=1}^ {\infty} \exp\left( -2^j \dfrac{n^ {2-\nu}}{64 \log(\dfrac{n}{n^ {1/4}  C})/4^ {\nu}} \alpha \right)  \leq  \exp\left( - \dfrac{n^{2-\nu}}{64 \log(\dfrac{n}{n^ {1/4}  C})/4^ {\nu}} \alpha \right).
\end{align}
Hence 
\begin{align}
 P(A_ {2, n} > \alpha) \lesssim C_3 \log n \sum_{j=1}^ {\infty}  \exp\left(- 2^j \dfrac{n^ {2-\nu}}{64 \log(\frac{n}{n^ {1/4}  C})/4^ {\nu}} \alpha \right) \lesssim  C_3 \log n \exp\left(- \frac{n^ {2-\nu}}{64 \log(\frac{n}{n^ {1/4} C})/4^\nu} \alpha \right),
\end{align}
for $\alpha \ge \alpha_n$.
\begin{align}
\int_{\alpha_n}^ {\infty}  P(A_ {2, n} > \alpha) d \alpha
\nonumber & \leq \int_ {\alpha_n}^ {\infty}  C_3 \log n \exp\left(- \dfrac{n^{2-\nu}}{64 \log(\frac{n}{n^ {1/4}  C})/4^ {\nu}} \alpha \right) d \alpha 
\\
\nonumber & \leq 64 C_3 \log n \frac{\log(\frac{n}{n^ {1/4} C})}{n^ {2-\nu}} \exp\left(- \dfrac{n^ {2-\nu}}{64 \log(\frac{n}{n^ {1/4} C})/4^ {\nu}} \times \frac{8}{n} \left(\frac{n C}{\log \frac{n}{n^ {1/4} C}} \right) \right).
\end{align}

\medskip

Let 
\begin{equation}
 n  C < \log(\dfrac{n}{n^ {1/4} C})(n 2^j \alpha/4)^ {\nu} \Longrightarrow   \alpha <  \frac{8}{n}\left(\frac{n C}{\log(\frac{n}{n^ {1/4}C})^{\nu}}\right)^ {1/\nu} \coloneqq \alpha_{n},
\end{equation}
Thus,
\[ -\dfrac{(n 2^j \alpha)^2 /16}{n C  + \log(\dfrac{n}{n^ {1/4} C})(n 2^j \alpha/4)^ {\nu}} \leq -\frac{(n 2^j \alpha)^2 /16}{2n C}.   \]
We can see that  asymptotically $\alpha_n > 1, ~ 2^j > 1$.
Thus, 

\medskip

\textbf{Step2}: $1 \leq  \alpha \leq  \alpha_{n}$ 

\medskip

Thus we have 

\medskip

\begin{align}
\nonumber & \sum_{j=1}^ {\infty}  \exp\left( 2 \frac{2^j \alpha}{\lambda_n}(L_n + 1) \log \left(\frac{ (L_n + 1)(N_n + 1) B_n}{ \dfrac{2^j
\alpha}{8 K_n} - \tau_n (L_n + 1)((N_n + 1) B_n)^{L_n +1}} \right)  - \dfrac{(n 2^j \alpha)^2 /16}{n C + \log(\frac{n}{n^ {1/4} C})(n 2^j \alpha/4)^ {\nu}} \right) 
\\
&  \leq \sum_{j=1}^ {\infty} \exp \left( 2 \dfrac{2^j \alpha}{\lambda_n}(L_n + 1) \log \left(16 K_n (L_n + 1)(N_n + 1) B_n  \right)  -\frac{(n 2^j \alpha)^2 /16}{2 n C} \right).
\end{align}
For $1 \leq \alpha \leq \alpha_{n}$,  we have $2^j \alpha \leq (2^j \alpha)^2$
\begin{align}
\nonumber & \sum_{j=1}^ {\infty} P\left\{\underset{ g \in \mathcal{G}_ {n, j, \alpha}}{\sup} \left[\E [g (Z_0)] - \dfrac{1}{n} \sum_{i=1}^ {n} g (Z_i) \right] >  \frac{2^j \alpha}{2} \right\} 
\\
& \leq C_3 \log n \sum_{j=1}^ {\infty} \exp\left( 2^j \alpha \left(\frac{2}{\lambda_n}(L_n + 1) \log \left(16 K_n (L_n + 1)(N_n + 1) B_n  \right)  -\frac{n^2}{32 n C} \right) \right).
\end{align}
We can see that for n large enough
\[  \delta_{n, 2} \coloneqq \dfrac{2}{\lambda_n}(L_n + 1)\log\left(16 K_n (L_n + 1)(N_n + 1) B_n  \right)  \leq \frac{n^2 /16}{4 n C}.  \] 
Thus,
\begin{align}
\sum_{j=1}^ {\infty} P\Big\{\underset{ g \in \mathcal{G}_ {n, j, \alpha}}{\sup} \left[\E [g (Z_0)] - \dfrac{1}{n} \sum_{i=1}^{n} g (Z_i) \right] >  \frac{2^j \alpha}{2} \Big\}
\leq C_3 \log n \sum_{j=1}^ {\infty} \exp\left( -2^j\dfrac{n}{64 C} \alpha \right).
\end{align}
Set $\beta_n^{'} = \dfrac{n}{64 C} $  using similar arguments as (\ref{inequ1}) we have 
\[P(A_ {2, n} > \alpha) \lesssim C_3 \log n \sum_{j=1}^ {\infty} \exp \left( -2^j \dfrac{n}{64 C} \alpha \right) \lesssim C_3 \log n \exp\left( - \dfrac{n}{64 C} \alpha \right),  \]
for $1 \leq \alpha \leq \alpha_{n}$. 
Which implies  
\begin{align}
\nonumber \int_{1}^ {\alpha_n} P(A_ {2, n} > \alpha)  & \leq C_3 \log n \int_{1}^ {\alpha_n}  \exp\left( -\dfrac{n}{64 C}\alpha \right) d \alpha  
\\
\nonumber & \leq \frac{64 C C_3\log n}{n}\exp\left(-\dfrac{n}{64 C} \times \dfrac{8}{n}\left(\frac{n C}{\log(\frac{n}{n^ {1/4} C})} \right)^{1 /\nu} \right) - \frac{64 C C_3 \log n}{n} \exp\left(-\dfrac{n}{64 C} \right).
\end{align}
\textbf{Step}3: $0 < \alpha\leq 1 < \alpha_n$.

Recall the    the assumption  
\[\tau_n \leq \dfrac{ \beta_n }{16  K_n  (L_n + 1)((N_n + 1) B_n)^{L_n +1}},  \]
with  the conditions $ \beta_n := (\log n)^{\nu_5}/n^{\nu_6}$  for some  $\nu_5, \nu_6 >0$
and
\begin{equation}\label{assump_beta}
\Big( \nu_6 < 1/2, \nu_4 + \nu_6 <1  \Big) \text{ or } \Big( \nu_6 < 1/2, \nu_4 + \nu_6 =1, \nu_5 > 1-\nu_3 \Big).
\end{equation}
We have
for  all $\alpha \in (\beta_n, 1)$, 
\begin{align}
\nonumber & \sum_{j=1}^\infty \exp \left( 2 \frac{2^j \alpha}{ \lambda_n}(L_n + 1)\log \left(\frac{ (L_n + 1)(N_n + 1)B_n}{ \dfrac{2^j \alpha}{8 K_n} - \tau_n (L_n + 1)((N_n + 1) B_n)^{L_n +1} } \right)  -\frac{(n 2^j \alpha)^2/16}{n C + \log (\frac{n}{n^{1/4} C})(n 2^j \alpha/8)^\nu} \right)
\\
\nonumber &\leq \sum_{j=1}^\infty \exp\left( \frac{2(L_{n} + 1 )2^{j}\alpha}{\lambda_{n}}\log\left(\frac{ 16 K_{n}(L_{n} + 1)(N_{n} + 1)B_{n}}{\beta_n}\right)  -\frac{(n2^{j}\alpha)^{2}}{32 n C}\right) 
\\
\nonumber & \leq \sum_{j=1}^{\infty} \exp\left( \frac{2(L_{n} + 1 )(2^{j})^2\alpha}{\lambda_{n}}\log\left(\frac{16 K_{n}(L_{n} + 1)(N_{n} + 1)B_{n}}{\beta_n}\right)  -\frac{(n2^{j}\alpha)^{2}}{32 n C}\right) 
\\
 \nonumber  & \leq \sum_{j=1}^{\infty} \exp\Bigg[ - (2^j)^2 \Bigg( \frac{n \alpha^{2}}{ 32 C } -  \frac{2(L_n + 1 )\alpha}{\lambda_n}\log\left(\frac{16 K_n (L_n + 1)(N_n + 1)B_n}{\beta_n}\right)  \Bigg) \Bigg] \\
\label{ineq_exp_phi_n} & \leq \sum_{j=1}^{\infty} \exp\big(- 4^j \phi_n (\alpha) \big)
\end{align}
with 
\begin{equation}\label{def_phi_n}
 \phi_n(\alpha) \coloneqq  \frac{n \alpha^{2}}{32 C} -  \frac{2(L_n + 1 )\alpha}{\lambda_n}\log\left(\frac{16 K_n (L_n + 1)(N_n + 1)B_n}{\beta_n}\right)  ~ \text{ for all}  ~ \beta_n < \alpha\leq 1 < \alpha_n.
\end{equation}
Since $K_n, L_n, N_n, B_n \geq 1$, we have $16 K_n (L_n + 1)(N_n + 1)B_n / \beta_n >1$ and therefore,
\begin{equation}\label{eq_phi_n_alpha}
\phi_n (\alpha) \geq  \frac{n \beta_n^ 2}{32 C} -  \frac{2(L_n + 1 )\beta_n}{\lambda_n}\log\left(\frac{16 K_n (L_n + 1)(N_n + 1)B_n}{\beta_n}\right) = \phi_n (\beta_n), ~   \text{ for all} ~ \beta_n < \alpha\leq 1 < \alpha_n. 
 \end{equation}
With the assumptions $\lambda_n \asymp (\log n)^{\nu_3}/ n^{\nu_4}$ for some $0< \nu_4 < 1$, $\nu_6 < 1/2$ and (\ref{assump_beta}), we get,
\begin{equation}\label{def_varphi_n}
\varphi_n \coloneqq  \phi_n (\beta_n) \limiten \infty .
\end{equation} 
One can see that,
\begin{equation*}
\exp(-\beta 4^j) \leq \exp(-\beta) \times \exp(-\beta j), \text{for all} ~  \beta \geq 0, ~ j \in \N. 
\end{equation*} 
Hence, it follows from (\ref{ineq_exp_phi_n}), (\ref{eq_phi_n_alpha}) and (\ref{def_varphi_n}) that, for all $\alpha \in (\beta_n, 1)$, and for $n$ large enough,
\begin{align}
\nonumber &\sum_{j=1}^ {\infty}\exp\left( 2\frac{2^j \alpha}{\lambda_n}(L_n + 1)\log\left(\frac{16 K_n (L_n + 1)(N_n + 1)B_n}{2^j \alpha} \right)  -\frac{(n 2^j \alpha)^ {2} /16}{n C + \log(\frac{n}{n^ {1/4} C})(n 2^j \alpha /8)^ {\nu}} \right)  \\
\nonumber & \leq \sum_{j=1}^ {\infty} \exp\big(- 4^j \varphi_n \big)
\leq \sum_{j=1}^ {\infty} \exp(-\varphi_n) \exp(-\varphi_n j)
\leq \dfrac{ \exp(-2 \varphi_n)}{ 1 - \exp(-\varphi_n) } \leq 2 \exp(-2 \varphi_n). 
\end{align} 
Therefore, for sufficiently large $n$,
\[  \int_0^1 P(A_ {2,n} > \alpha) d \alpha \leq \beta_n +  \int_{\beta_n}^1 P(A_ {2,n} > \alpha) d \alpha  \leq   \beta_n + 2 \exp(-2 \varphi_n) \leq 2 \beta_n,  \]
where $\beta_n$ is defined above, and satisfies (\ref{assump_beta}).
Thus, for $\nu_5 >0, \; \nu_6 < 1/2$, we have
\begin{align}
\nonumber \E[A_{2, n}] & = \int_{0}^ {\infty} P (A_{2, n} > \alpha) d \alpha 
\\
\nonumber & \leq \int_{0}^ {\beta_n} P (A_{2, n} > \alpha) d \alpha  +  \int_{\beta_n}^1 P (A_ {2, n} > \alpha) d \alpha + \int_{1}^ {\alpha_n} P(A_ {2, n} > \alpha) d \alpha +  \int_{\alpha_n}^ {\infty} P (A_{2, n} > \alpha) d \alpha
\\
\nonumber & \leq 2\beta_n + \frac{64 C C_3\log n}{n}\exp\left(-\frac{n}{64 C}\times \frac{8}{n}\left(\frac{n C}{\log(\frac{n}{n^ {1/4}C})}\right)^{1/\nu}\right)- \frac{64 C C_3 \log n}{n}\exp\left(-\frac{n}{64 C}\right)  
\\
 \nonumber & +  32 C_3 \log n \frac{\log(\frac{n}{n^ {1/4} C})} {n^ {2-\nu}}\exp\left(-\frac{n^ {2-\nu}}{32\log(\frac{n}{n^ {1/4} C}) /8^ {\nu}} \times \frac{8}{n} \left(\frac{n C}{\log\frac{n}{n^ {1/4} C}} \right) \right)
 \\
&  \lesssim 2\beta_n.
\end{align}

For $A_ {4, n} $ we choose a neural network function $h_ {n} ^0 \in \mathcal{H}_{\sigma}(L_n, N_n, B_n, F) $ such that 
\[ \| h_ {n}^ {0} - h^ {*} \|_{2, P_ {X_0}}^ {2} + J_{\lambda_n, \tau_n}(h_ n ^0) \leq \underset{h \in \mathcal{H}_{\sigma}(L_n, N_n, B_n, F)}{\inf}\Big[\|h- h^ {*} \|_{2, P_ {X_0}} ^2 +  J_{\lambda_n, \tau_n} (h) \Big] + n^ {-1}.  \] 
Then by the basic inequality 
\[ \frac{1}{n} \sum_{i=1}^n \left(Y_{i} - \widehat{h}_n (X_i) \right) ^2  + J_{\lambda_n, \tau_n} (\widehat{h}_n) \leq  \frac{1}{n} \sum_{i=1}^n \left(Y_{i} - h (X_{i}) \right)^2  + J_{\lambda_n, \tau_n} (h),  \]
 for any  $ h \in \mathcal{H}_ {\sigma}(L_n, N_n, B_n, F) $, we have  
\begin{align}
 A_ {4, n} \nonumber & = 2 \Big[ \frac{1}{n} \sum_{i=1}^n (Y_i - \widehat{h}_n (X_i) )^2 - \frac{1}{n} \sum_{i=1}^n (Y_{i} - h_n^0 (X_i) )^2 \Big] +  2 J_{\lambda_n, \tau_n} (\widehat{h}_n)  
 \\
 &  +  2 \Big[ \frac{1}{n} \sum_{i=1}^n (Y_i - h_n^0 (X_i) )^2 - \frac{1}{n} \sum_{i=1}^n (Y_i - h^ {*} (X_i) )^2  \Big]
\\
 \nonumber &   \leq  2J_{\lambda_n, \tau_n} (h_n^0 ) + 2 \Big[ \frac{1}{n} \sum_{i=1}^n (Y_i - h_n^0 (X_i))^2 - \frac{1}{n} \sum_{i=1}^n (Y_i - h^ {*}(X_i) )^2  \Big]
\end{align}
and we have
\begin{align}
\E[ A_ {4, n}] \nonumber \leq  2J_{\lambda_n, \tau_n} (h_n^0 ) + \frac{2}{n} \sum_{i=1}^n \Big[ \E (Y_i - h_n^0 (X_i)) ^2 - \E (Y_i - h^ {*} (X_i)) ^2  \Big],
\end{align}
One can see that for $i=1, \dots, n$,
\[  \E[(Y_i - h_n^0 (X_i)) ^2] - \E[(Y_i - h^ {*}(X_i)) ^2] \leq \| {h}_n^0 - h^ {*} \|_{2, P_{X_0}} ^2.  \]
Thus,
\begin{align*}
\E[ A_ {4, n}] \nonumber  & \leq  2 J_{\lambda_n, \tau_n} (h_n^0) + 2 \| {h}_n^0 -h^{*} \|_{2, P_{X_0}} ^2 &  \leq  2 \underset{h \in \mathcal{H}_{\sigma} (L_n, N_n, B_n, F)}{\inf} \Big[ \|h- h^ {*} \|_{2, P_{X_0}} ^2 + J_{\lambda_n, \tau_n} (h) \Big] + 2 n^ {-1}.
\end{align*}
Thus, 
\begin{align}
\nonumber & \E \Big[ \| \widehat{h}_n - h^ {*}\|_ {2, P_ {X_0}}^ {2} \Big]   = \E \left[ \sum_{i=1} ^4 A_ {i, n} \right] 
 \lesssim 3 \dfrac{\log n}{n} + 2\dfrac{(\log n)^ {\nu_5}}{n^ {\nu_6}} +  2 \underset{h \in \mathcal{H}_ {\sigma} (L_n, N_n, B_n, F)}{\inf} \Big[ \| h- h^ {*} \|_ {2, P_ {X_0}} ^2  \Big] + 2 n^ {-1} 
\\
& \lesssim  2 \Bigg( \underset{h \in \mathcal{H}_ {\sigma}(L_n, N_n, B_n,  F)}{\inf} \{ \| h - h^ {*}\|_ {2, P_ {X_0}} ^2 + \lambda_n \| \theta (h) \|_ { \text{clip}, \tau_n} \} \lor  \dfrac{(\log n)^ {\nu_5}}{n^ {\nu_6}} \Bigg). 
\end{align}
This completes the proof of the theorem.                             

\qed  

\medskip

\subsection{Proof of Theorem \ref{thm2}}   
Let $\epsilon_{n}= n^ {-\nu_4 /(\kappa + 2)} $. From the Theorem (\ref{thm1}),  the assumption (\ref{equ_assump_thm2}) and the fact that $\|\theta (h) \|_ {\text{clip}, \tau} \leq \|\theta (h)\|_0$ for any $\tau > 0$, we have that for any $ h^ {*} \in \mathcal{H}^ {*} $, 
\begin{align}\label{conver_rate}
\nonumber   & \underset{h \in \mathcal{H}_ {\rho, H^ {*}}: h^ {*} \in \mathcal{H}^ {*}}{\sup} \E \Big[ \| \widehat{h}_n - h^ {*} \|_ {2, P_ {X_0}}^ {2} \Big] 
\\
\nonumber & \lesssim \underset{h \in \mathcal{H}_ {\rho, H^ {*}}: h^ {*} \in \mathcal{H}^ {*}}{\sup} ~ \underset{h \in \mathcal{H}_ {\sigma}(L_n, N_n, B_n,  F, C \epsilon_n^ {-\kappa} (\log n) ^r )}{\inf}  \Big\{ \| h - h^ {*}\|_ {2, P_ {X_0}}^ {2} + \lambda_n \| \theta (h) \|_ { \text{clip}, \tau_n}  \Big\} \lor  \dfrac{(\log n)^ {\nu_5}}{n^ {\nu_6}} 
\\
\nonumber  & \lesssim  \underset{h \in \mathcal{H}_ {\rho, H^ {*}}: h^ {*} \in \mathcal{H}^ {*}}{\sup} \Big[ ~ \underset{h \in \mathcal{H}_ {\sigma} (L_n, N_n, B_n,  F,  C\epsilon_n^ {-\kappa} (\log n) ^r)} {\inf}  \| h - h^ {*} \|_ {2, P_ {X_0}}^ {2}   + \lambda_n \epsilon_n^ {-\kappa} (\log n)^r \Big] \lor  \dfrac{(\log n)^ {\nu_5}}{n^ {\nu_6}}
\\
\nonumber & \lesssim  \underset{h \in \mathcal{H}_{\rho, H^ {*}}: h^ {*} \in \mathcal{H}^ {*}}{\sup} \Big[ \epsilon_n^ {2} +  \dfrac{(\log n)^{\nu_3}}{n^ {\nu_4}}   \epsilon_n^ {-\kappa} (\log n) ^r \Big] \lor  \dfrac{(\log n)^ 
 {\nu_5}}{n^ {\nu_6}} 
\\
 & \lesssim \dfrac{(\log n)^ {r + \nu_3}}{n^{2\nu_4 /(\kappa + 2)}} \lor   \dfrac{(\log n)^ {\nu_5}}{n^ {\nu_6}}.
\end{align}
Thus, the theorem follows.
\qed 

\subsection{Proof of Theorem \ref{thm3}}
%
%
Let us decompose $ \mathcal{E}_ {Z_0} (\widehat{h}_n) $ as as follows.
\begin{equation}
\mathcal{E}_{Z_0} (\widehat{h}_n)  = \E [ \ell (Y_0 \widehat{h}_n (X_0)) ] -  \E [ \ell (Y_0 h^{*}_{\ell} (X_0)) ]  \coloneqq B_ {1, n} + B_ {2, n},
\end{equation}
where 
\begin{align}
\nonumber B_ {1, n} & =  \Big[  \E [ \ell (Y_0 \widehat{h}_n (X_0)) ] -  \E [ \ell (Y_0 h^{*}_{\ell} (X_0)) ] \Big]  -  \dfrac{2}{n} \sum_{i=1}^n   \Big[  \ell(Y_i \widehat{h}_n (X_i)) -  \ell(Y_i h^{*}_{\ell} (X_i)) \Big] - 2 J_{\lambda_n, \tau_n}(\widehat{h}_n ); 
\\
\nonumber  B_ {2, n} & = \dfrac{2}{n} \sum_{i=1}^n  \Big[  \ell (Y_i \widehat{h}_n (X_i)) -  \ell (Y_i h^{*}_{\ell} (X_i)) \Big] + 2 J_{\lambda_n, \tau_n}(\widehat{h}_n). 
\end{align}
One can bound $ B_{1, n} $ in the same way as for $ A_{2, n} $ in the proof of Theorem \ref{thm1}. Let
$ \Delta (h) (Z_0) \coloneqq  \ell (Y_0  h (X_0)) - \ell (Y_0  h^*_\ell (X_0)) $, with $ Z_0 \coloneqq (X_0, Y_0) $. 
Set
\begin{equation*}
\mathcal{H}_ {n, j, \alpha} \coloneqq \{ h \in  \mathcal{H}_ {\sigma} (L_n, N_n, B_n, F): 2^ {j -1} \ind{( j \ne 0)} \alpha \leq J_ {\lambda_n, \tau_n} (h) \leq 2^j \alpha  \},
\end{equation*}
and 
\begin{equation}
\mathcal{G}_{n, j, \alpha} \coloneqq \Big\{ \Delta (h): \R^d \times \{ -1, 1 \} \mapsto \R: h \in \mathcal{H}_{n, j, \alpha} \Big\}.
\end{equation}
It holds from the definition of $h^{*}_\ell$ that $ \E \Delta (h) (Z_0) \ge 0$.
 We have for all $\alpha > 0$,
\begin{align}
\nonumber P ( B_ {1, n} > \alpha)  & \leq  P\Big( \underset{ h \in \mathcal{H}_ {\sigma} (L_n, N_n, B_n, F)}{ \sup} \dfrac{\E \Delta (h) (Z_0) - \dfrac{1}{n} \sum_{i=1}^n \Delta (h) (Z_i)}{ \alpha 
 + 2 J_ {\lambda_n, \tau_n} (h) + \E \Delta (h) (Z_0)} \ge \dfrac{1}{2}  \Big)
 \\
\nonumber & \leq \sum_{j = 0}^ {\infty} P \Big(\underset{ h \in \mathcal{H}_ {n, j, \alpha}}{ \sup} \dfrac{ \E \Delta (h) (Z_0) - \dfrac{1}{n} \sum_{i=1}^n \Delta (h)(Z_i)}{
  2^j \alpha + \E \Delta (h) (Z_0)}  \ge  \dfrac{1}{2}  \Big)\\
 \nonumber & \leq  \sum_{j = 0}^ {\infty} P \Big(\underset{ g \in \mathcal{G}_ {n, j, \alpha}}{ \sup} \dfrac{ \E g (Z_0) - \dfrac{1}{n} \sum_{i=1}^n g (Z_i)}{  2^j \alpha }  \ge 
 \dfrac{1}{2}  \Big) 
  \leq  \sum_{j = 0}^ {\infty} P \Big(\underset{ g \in \mathcal{G}_ {n, j, \alpha}}{ \sup}  \E g (Z_0) - \dfrac{1}{n} \sum_{i=1}^n g (Z_i)\ge 
 \dfrac{  2^j \alpha   }{2}  \Big) 
 \\
\nonumber &  \leq  \sum_{j = 0}^ {\infty} P \Big( \underset{ g \in \mathcal{G}_ {n, j, \alpha}}{ \sup}  \sum_{i=1}^n \Big (\E g (Z_0) -   g (Z_i) \Big) \ge 
 \dfrac{2^j \alpha n} {2}  \Big). 
\end{align} 
Let $ \Delta (h_1), ~ \Delta (h_2)  \in  \mathcal{G}_ {n, j, \alpha}$ and $(x, y) \in  \R^d \times \{-1, 1\} $, we have 
\begin{align*}
\| \Delta (h_1) (x, y) - \Delta (h_2) (x, y) \|_{\infty}  & \leq \| \ell (y h_1 (x) ) -  \ell ( y h_2 (x) )  \|_{\infty} 
\leq \mathcal{K}_{\ell} \vert h_1 (x) - h_2 (x) \vert.
\end{align*}
Thus,
\[ \mathcal{ N} \Big( \varepsilon, \mathcal{G}_{n, j, \alpha}, \| \cdot \|_{\infty}  \Big) 
 \leq  \mathcal{ N} \Big( \dfrac{\varepsilon}{\mathcal{K}_{\ell}}, \mathcal{H}_{n, j, \alpha}, \| \cdot \|_{\infty}  \Big).  \]
As stressed in the proof of Theorem \ref{thm1}, the process $ \Big (g(Z_t) \coloneqq \ell( h(X_t), Y_t) \Big)_{ t \in \Z} $ is also $\psi$-weakly dependent.
From  (\ref{inclusion}) and  (\ref{cover_number}) we have,
\begin{align}\label{cover_number_equ2}
\mathcal{N}( \varepsilon, \mathcal{G}_{n, j, \alpha}, \| \cdot \|_\infty)  \nonumber & \leq \mathcal{N} \Big( \dfrac{\varepsilon}{ \mathcal{K}_{\ell}}, \mathcal{H}_{n, j, \alpha}, \| \cdot \|_\infty  \Big) 
\\
\nonumber & \leq \mathcal{N} \Big(\dfrac{ \varepsilon}{\mathcal{K}_{ \ell}}, \mathcal{H}_{\sigma}(L_{n},N_{n}, B_{n}, F, \frac{2^j \alpha}{\lambda_n}),  \| \cdot \|_\infty  \Big)
\\
& \leq  \exp \Big(2 \dfrac{2^j \alpha}{\lambda_n}(L_n + 1) \log \Big(\dfrac{(L_n + 1)(N_n + 1)B_n}{\varepsilon/ \mathcal{K}_{\ell} - \tau_n (L_n +1)((N_n + 1)B_n)^{L_n + 1}} \Big) 
 \Big).
\end{align} 
Let $ \varepsilon \coloneqq  \dfrac{2^j \alpha}{2} $.
By using similar arguments as in (\ref{equ_5_18}).
We have
\begin{align}
\nonumber  & \sum_{j=1}^\infty P\Big\{\underset{ g \in \mathcal{G}_ {n, j, \alpha}}{\sup} \left[\E [g (Z_0)] - \dfrac{1}{n} \sum_{i=1}^n g (Z_i) \right] >  \frac{2^j \alpha}{2} \Big\} 
\\
\nonumber &  \leq C_{3}  \exp( \log \log n)  \\
\nonumber & \times \sum_{j=1}^\infty \exp\left( 2 \frac{2^j \alpha}{\lambda_n} (L_n + 1) \log \left(\frac{ (L_n + 1)(N_n + 1) B_n}{ \dfrac{2^j \alpha}{2 \mathcal{K}_{\ell}} - \tau_n(L_n +1)((N_n + 1)B_n)^{L_n + 1}} \right)  -\frac{(n 2^j \alpha)^2/16}{n C + \log(\frac{n}{n^ {1/4} C})(n 2^j \alpha /8)^ {\nu}} \right).
\end{align}
With the assumptions $ L_n \lesssim \log n, ~  N_n \lesssim n^{\nu_1}, ~ B_n \lesssim n^ {\nu_2}, ~ \tau_n \leq \dfrac{\beta_n}{4 \mathcal{K}_{\ell} (L_n +1)((N_n + 1) B_n)^ {L_n + 1}} $; under the conditions on $\nu_1 >0, ~ \nu_2 >0, ~ \beta_n, \nu_5, ~ \nu_6 $, by going as in $A_ {2, n} $ (see the proof of Theorem \ref{thm1}), we get,
\[ \E [B_ {1,n}] \lesssim  2 \dfrac{( \log n)^ { \nu_5} }{ n^{\nu_6}}. \]

\medskip

Let us deal now with $B_ {2, n} $.
We choose a neural network function $h_n^ {\circ} \in  \mathcal{H}_ {\sigma} (L_n, N_n, B_n, F) $ such that
\begin{equation}
\mathcal{E}_{Z_0} (h_n^\circ) + J_{\lambda_n, \tau_n} (h_n^\circ) \leq \underset{ h \in \mathcal{H}_ {\sigma} (L_n, N_n, B_n, F) }{ \inf} \Big[ \mathcal{E}_{Z_0} (h) +  J_ {\lambda_n, \tau_n} (h) \Big] + \dfrac{1}{n}.
\end{equation}
Then, from the basic inequality, 
\begin{equation*}
\dfrac{1}{n} \sum_{i=1}^{n} \ell(Y_i \widehat{h}_n (X_i)) +  J_{\lambda_n, \tau_n}(\widehat{h}_n) \leq \dfrac{1}{n} \sum_{i=1}^n  \ell(Y_i h (X_i))  +    J_{\lambda_n, \tau_n}(h), 
\end{equation*}
we have,
\begin{align*}
\nonumber B_{2, n} & = \dfrac{2}{n} \sum_{i=1}^n  \Big[  \ell(Y_i \widehat{h}_n (X_i)) -  \ell(Y_i h_n^\circ (X_i))  \Big] + 2 J_{\lambda_n, \tau_n}(\widehat{h}_n ))
+ \dfrac{2}{n} \Big[ \ell(Y_i h_n^\circ (X_i)) - \ell(Y_i h^{*}_{\ell} (X_i)) \Big]
\\
\nonumber & \leq  2 J_{\lambda_n, \tau_n}(h_n^\circ ) +  \dfrac{2}{n} \Big[ \ell(Y_i h_n^\circ (X_i)) - \ell(Y_i h^{*}_{\ell} (X_i)) \Big]
\\
& \leq 2 \underset{ h \in \mathcal{H}_{\sigma} (L_n, N_n, B_n, F) }{ \inf} \Big[ \mathcal{E}_{Z_0} (h) +  J_{\lambda_n, \tau_n}(h) \Big] + \dfrac{1}{n}.
\end{align*}
Hence
\begin{equation*}
\E \Big[ \mathcal{E}_{Z_0} (\widehat{h}_n) \Big] \leq 2 ~ \underset{ h \in \mathcal{H}_{\sigma} (L_n, N_n, B_n, F) }{ \inf} \Big\{ \mathcal{E}_{Z_0} (h) +  \lambda_n \| \theta(h) \|_{\text{clip}, \tau_n} \Big\} \lor  \dfrac{ (\log n)^{\nu_5}}{ n^{\nu_6}}.
\end{equation*}
This establishes the theorem.
\qed 
\subsection { Proof of Theorem \ref{thm4}}
Under the assumption $\ell ~ \text{is} ~ \mathcal{K}_{\ell}-\text{Lipschitz} $, we have
\
\begin{align*}
|\mathcal{E}_{z_0} (h) | & =  | \E[\ell(Y_0 h (X_0))] - \E[\ell(Y_0 h_{\ell}^{*} (X_0))]|  
\\
& \leq \E[|\ell (h (X_0)) - \ell (h_{\ell}^{*} (X_0))|] 
\\
&  \leq K_{\ell} \E [|h (X_0) - h_{\ell}^{*} (X_0)|]
\end{align*}
Thus,
\[\mathcal{E}_{Z_0} (h) \leq \mathcal{K}_{\ell} \| h - h_{\ell}^{*} \|_{1, P_{X_0} }. 
  \]
 Let $ \epsilon_n = n^{-\dfrac{\nu_4}{\kappa + 1}}$. By Theorem \ref{thm3}, the condition (\ref{equ_assump_thm4}), and the fact that $ \| \theta (h)\|_{\text{clip}, \tau} \leq \| \theta\|_0$ for any  $ \tau > 0$, we have that for any $ h_{\ell}^ {*} \in \mathcal{H}^ {*} $,
\begin{align}
\nonumber \E \Big[ \mathcal{E}_{Z_0} (\widehat{h}_n)\Big] & \lesssim \underset{h \in \mathcal{H}_{\sigma}(L_n, N_n, B_n, F, C \epsilon_n^{-\kappa} (\log n)^r)}{\inf} \Big\{ \mathcal{ E}_{Z_0} (h) +  \lambda_n \| \theta (h) \|_{\text{clip}, \tau_n} \Big\} \lor \dfrac{ (\log n)^{\nu_5}}{n^{\nu_6}}
\\
\nonumber & \lesssim  \Big[  \underset{h \in \mathcal{H}_{\sigma}(L_n, N_n, B_n, F, C \epsilon_n^ {-\kappa} (\log n) ^r)}{\inf} \| h -  h_{\ell}^ {*} \|_ {1, P_ {X_0}}  + \lambda_n \epsilon_n^ {- \kappa}  (\log n) ^r \Big] \lor \dfrac{ (\log n) ^{\nu_5}}{n^ {\nu_6}}
\\
\nonumber & \lesssim \dfrac{ (\log n)^ {r + \nu_3}}{n^{ \dfrac{\nu_4}{\kappa +1}}} \lor \dfrac{ (\log n)^ {\nu_5} }{n^ {\nu_6}}.
\end{align}
Thus,
\begin{align}
\underset{H \in \mathcal{Q}_{H}^ {*}: h_{\ell}^ {*} \in \mathcal{H}^ {*}}{\sup} \E \Big[ \mathcal{E}_{Z_0} (\widehat{h}_n) \Big] \lesssim \dfrac{ (\log n)^ {r + \nu_3}} {n^ { \dfrac{\nu_4}{\kappa + 1}}} \lor  \dfrac{ (\log n)^ {\nu_5} }{n^ {\nu_6}}.
\end{align}

\qed

\end{document}